\documentclass[runningheads]{llncs}

\usepackage{eccv}

\usepackage{eccvabbrv}

\usepackage{graphicx}
\usepackage{booktabs}
\usepackage{wrapfig}

\usepackage[accsupp]{axessibility}  %

\usepackage{hyperref}

\usepackage{orcidlink}

\newcommand{\shortname}{Deform360\xspace}
\usepackage{lipsum} 
\usepackage{tabularx}
\usepackage{adjustbox}

\usepackage{pifont}%
\newcommand{\cmark}{\textcolor{ForestGreen}{\ding{51}}}%
\newcommand{\xmark}{\textcolor{red}{\ding{55}}}%

\begin{document}

\title{\shortname: A Massive Multi-view Visuotactile Dataset for Deformable World Models} 

\titlerunning{Deform360}

\author{Hongyu Li\inst{1,2*}\orcidlink{0000-0002-4588-9929} \and
Wanjia Fu\inst{1*}\orcidlink{0000-0002-7210-3725} \and
Xiaoyan Cong\inst{1}\orcidlink{0009-0003-2546-694X} \and
Zekun Li\inst{1}\orcidlink{0000-0002-1461-4121} \and \\
Binghao Huang\inst{2}\orcidlink{0009-0008-1758-5531} \and
Hanxiao Jiang\inst{2}\orcidlink{0000-0001-6245-361X} \and
Xintong He\inst{1} \and
Yiqing Liang\inst{1}\orcidlink{0000-0001-9388-3647} \and \\
Rao Fu\inst{1}\orcidlink{0000-0002-0115-0831} \and
Tao Lu\inst{1}\orcidlink{0009-0000-8830-3820} \and
Srinath Sridhar\inst{1}\orcidlink{0000-0003-4663-3324} \and \\
Kevin A. Smith\inst{3}\orcidlink{0000-0001-5009-0460} \and 
George Konidaris\inst{1}\orcidlink{0000-0002-4460-2519} \and
Yunzhu Li\inst{2}\orcidlink{0000-0002-1111-2150}
}

\authorrunning{H. Li et al.}

\institute{
Brown University, Providence, RI \and
Columbia University, New York, NY \and
Massachusetts Institute of Technology, Cambridge, MA \\
$^*$ Equal contribution \\
}
\maketitle

\begin{abstract}
Predicting object dynamics (i.e., world modeling) is a fundamental challenge for robotic manipulation, and modeling deformable objects presents a particularly difficult case due to their high-dimensional state spaces and complex material properties. While current world models approach this through two distinct paradigms: learning the dynamics over the 2D pixel space or more explicit 3D geometric space. A systematic understanding of their relative strengths and limitations remains elusive due to the lack of diverse, large-scale real-world data. To address this, we present \shortname, a large-scale visuotactile dataset featuring 198 daily-life objects, 1,980 interaction sequences, and over 215 hours of observations from 41 surround-view cameras and bimanual tactile grippers to capture both global motion and contact-induced local deformations. Leveraging a novel markerless visuotactile 3D tracking pipeline to extract dense geometry and motion, we systematically evaluate current state-of-the-art world models, comparing 2D video models against 3D particle models. Finally, we provide a preliminary demonstration indicating the real-world applicability of our dataset by performing robot planning tasks on deformable objects. Our analysis reveals key insights into the trade-offs between structural priors and scalability, providing a solid benchmark for future research in generalizable deformable object-centric world modeling.
Project website: \url{https://deform360.lhy.xyz}
\end{abstract}

\begin{figure*}[h!]
    \includegraphics[width=\textwidth]{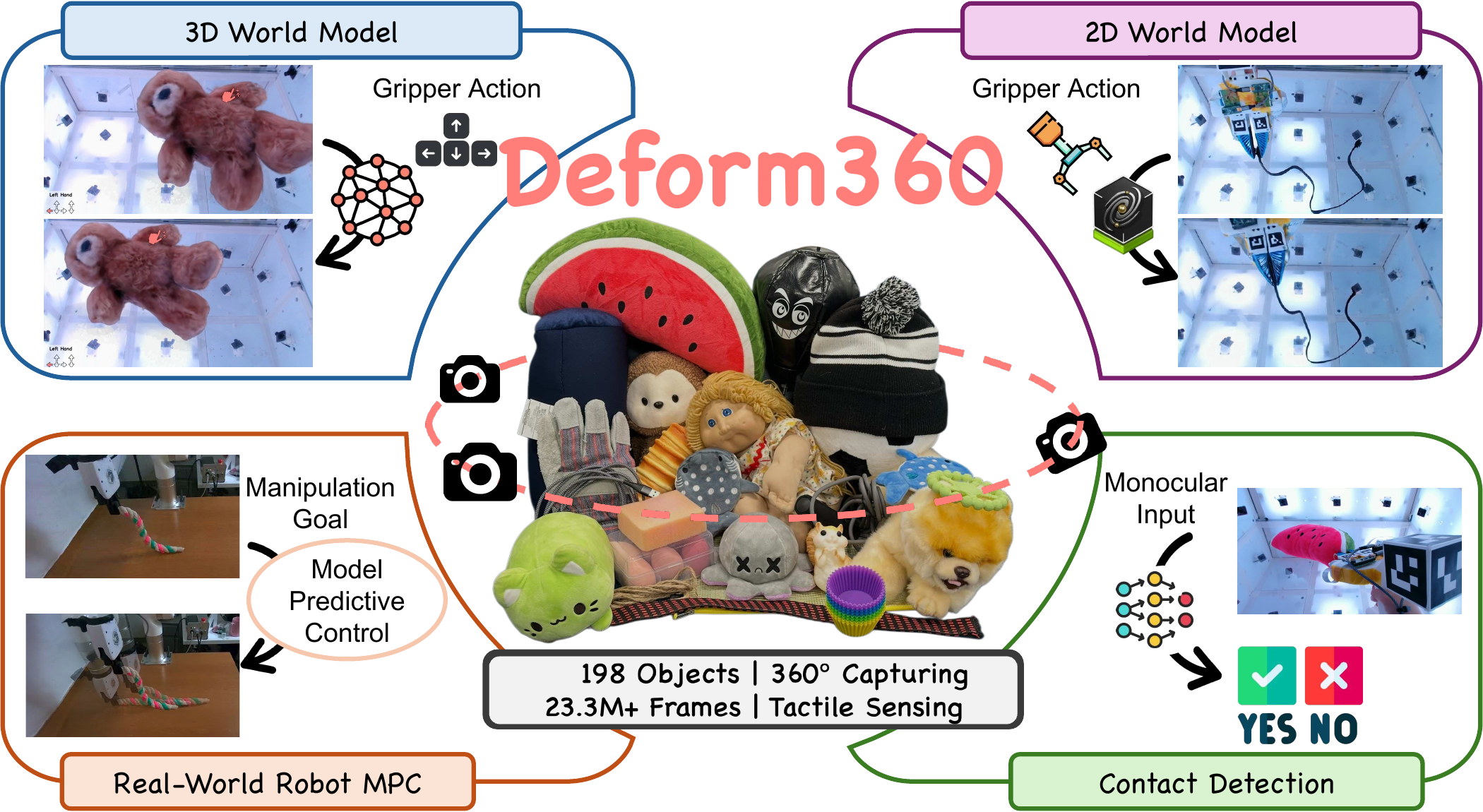}
    \caption{
    \textbf{Overview of \shortname}.
    We collect a massive multi-view visuotactile dataset with 198 deformable objects (a subset is shown), supporting 2D and 3D world models, contact detection, and real-world robot planning tasks.
    }
    \label{fig:teaser}
\end{figure*}

\section{Introduction}
\label{sec:introduction}

World modeling (\ie, action-conditioned future prediction) has emerged as a powerful paradigm for robots to simulate environments and plan actions~\cite{ai_review_2025, li_WorldModelBench_2025, kang_How_2024, bansal_VideoPhy_2024}. %
Despite recent progress, accurately predicting the behavior of deformable objects remains a significant challenge~\cite{cong20244drecons,duisterhof_DeformGS_2024, tang_Track_2022}.
Deformable bodies possess theoretically infinite degrees of freedom, and contact-induced local deformations are frequently occluded by end-effectors or the object itself~\cite{jin_Robust_2019,cong2025dytact,pokhariya_MANUS_2024}.
Consequently, capturing deformable dynamics accurately has attracted significant interest, particularly via tactile sensing to observe occluded interactions and provide ground-truth signals. %

Current models predicting deformable dynamics follow two paradigms: predicting dynamics over 2D pixel space (\eg, video generation)~\cite{nvidia_Cosmos_2025,wan_Wan_2025,kong_HunyuanVideo_2025,wang_Interactive_2026} or 3D geometric space~\cite{li_Learning_2019, zhang_ParticleGrid_2025, zhang_AdaptiGraph_2024, jiang_PhysTwin_2025a, chen_Differentiable_2024}.
Evaluating their relative strengths requires diverse, large-scale, multi-view real-world datasets, which are essential for verifying generalizability and extracting ground-truth 3D states.
However, existing datasets often lack object diversity~\cite{blanco-mulero_Benchmarking_2024, chen_Differentiable_2024, jiang_PhysTwin_2025a, obrist_PokeFlex_2025,yu2025generalizable}, rely on synthetic data~\cite{li_Learning_2019, li_Propagation_2019, zhou_ClothesNet_2023, bear_Physion_2021, tung_Physion_2023, kang_How_2024}, or miss high-fidelity annotations and contact-induced deformation modeling~\cite{coltraro_Tracking_2025,liang_Robo360_2023, li_Textureless_2025,li_WorldModelBench_2025}.

To address this, we present \shortname, a massive and diverse visuotactile dataset for large-scale deformable dynamics modeling.
The dataset encompasses \textbf{198} daily-use objects across \textbf{1,980} interaction sequences, spanning a wide spectrum of materials.
Our capture system features 41 surround-view cameras and bimanual tactile-equipped UMI grippers~\cite{chi_Universal_2024,zhu_Touch_2025}, ensuring $360^{\circ}$ observability of global dynamics and fine-grained contact-induced deformations.
Delivering over \textbf{215.7 hours} and \textbf{23.3 million frames} of visuotactile recordings, \shortname provides a foundational benchmark for 3D dynamics evaluation of world models.
Leveraging this data, we develop a novel markerless 3D tracking pipeline to generate high-fidelity particle annotations across varying materials.

Using these annotations, we formulate three core tasks to evaluate deformable world modeling:
1) contact prediction, demonstrating the utility of our synchronized tactile data by modeling the coupling between visual observations and physical contact events. %
2) world model benchmarking, providing a systematic evaluation of current state-of-the-art 2D video models~\cite{nvidia_Cosmos_2025} against 3D particle models~\cite{huang_ParticleFormer_2025,zhang_ParticleGrid_2025,jiang_PhysTwin_2025a} across multiple generalization settings.
3) robot planning, evaluating the real-world applicability of our dataset by deploying learned world models in a model predictive control (MPC) framework for deformable object manipulation.
Our evaluation reveals that while 3D particle models excel in low-data regimes due to structured physical priors, 2D video models leverage their scalability to achieve better generalization when provided with extensive data.

In summary, this paper makes four key contributions:
1) A large-scale, multi-view, visuotactile real-world dataset of deformable object interactions, including 198 diverse daily-life objects with 1,980 interactions from two tactile-equipped UMI grippers, captured using 41 surround-view cameras.
2) High-quality 3D reconstructions and tracking results for each episode, derived from a markerless multi-view visuotactile perception pipeline.
3) A systematic evaluation comparing state-of-the-art 2D video world models and 3D particle dynamics models on dynamics reconstruction and future prediction, revealing critical insights into their respective trade-offs in structural priors and scalability. %
4) A preliminary demonstration of real-world applicability through robot planning using MPC on deformable objects collected in \shortname.

\section{Related Work}
\label{sec:related_work}

\begin{table}[t!]
\renewcommand{\arraystretch}{1.0}
\setlength{\tabcolsep}{4pt}

\caption{
  \textbf{Comparison of real-world datasets containing deformable objects.} 
  We benchmark existing datasets based on key modalities for learning dynamics, including availability of 3D mesh data (Mesh), camera calibration (Calib.), \emph{markerless} (whether tracking does not rely on physical markers), inclusion of tactile data (Tactile), multi-view coverage (360$^{\circ}$ View), and number of camera views (Views). 
  We also report the scale of each dataset in terms of deformable objects (\# obj.), total frames (\# frames), number of distinct manipulation-primitive types (\# actions; see appendix), and number of object categories (\# cat.).
  `\cmark' indicates the presence of a specific modality, and `\xmark' indicates its absence.
}
\label{table:dataset}

\centering
\resizebox{\textwidth}{!}{%
\begin{tabular}{lcccccrrrrr}
\toprule
\textbf{Dataset}  & \textbf{Mesh}  & \textbf{Calib.} & \textbf{Markerless} & \textbf{Tactile} &  \textbf{360$^{\circ}$}  &  \textbf{Views} & \textbf{\# obj.} & \textbf{\# frames} & \textbf{\# actions} & \textbf{\# cat.} \\
\midrule
ClothSim2Real~\cite{blanco-mulero_Benchmarking_2024} & \cmark & \xmark & \cmark & \xmark & \xmark & 1 & 3 & $\sim$4.9k & 2 & 1 \\
ClothTrack~\cite{coltraro_Tracking_2025} & \xmark & \xmark & \xmark & \xmark & \xmark & 5 & 8 & $\sim$173k & 4 & 1 \\
DDER~\cite{chen_Differentiable_2024}  & \xmark  & \xmark & \xmark & \xmark & \xmark & 11 & 5 & $\sim$1.8M & 5  & 2  \\
Robo360~\cite{liang_Robo360_2023}  & \xmark & \cmark & \cmark & \xmark & \cmark & 86 & 17 & $\sim$2M & 7 & 7 \\
PokeFlex~\cite{obrist_PokeFlex_2025} & \cmark & \cmark & \cmark & \cmark & \cmark & 6 & 18 & 21.3k & 2 & 6 \\
DOT~\cite{li_Textureless_2025} & \cmark & \cmark & \xmark & \xmark & \cmark & 42 & 22 & $\sim$689k & 9 & 3 \\
HMDO~\cite{xie_HMDO_2023} & \cmark & \cmark & \cmark & \xmark & \cmark & 10 & 12 & 21.6k & 2 & 1 \\
HCOS~\cite{garcia-camacho_Household_2022} & \cmark & \xmark & \cmark & \xmark & \xmark & 1 & 27 & 64 & 3  & 1 \\
PhysTwin~\cite{jiang_PhysTwin_2025a} & \xmark & \cmark & \cmark & \xmark & \xmark & 3 & 11 & 7.2k & 4 & 4 \\
PGND~\cite{zhang_ParticleGrid_2025} & \cmark & \cmark & \cmark & \xmark & \cmark & 4 & 8 & $\sim$1M & 6 & 6 \\
\shortname (Ours) & \cmark & \cmark & \cmark & \cmark & \cmark & 41 & 198 & $\sim$23.3M & 13 & 17 \\
\bottomrule
\end{tabular}%
}
\end{table}

\subsection{Deformable Dataset}
Existing datasets for deformable objects often focus on specific categories or lack essential modalities for comprehensive world modeling (\cref{table:dataset}). 
Early benchmarks primarily targeted thin-shell objects such as cloth~\cite{blanco-mulero_Benchmarking_2024, coltraro_Tracking_2025, garcia-camacho_Household_2022} or linear objects like ropes and cables~\cite{chen_Differentiable_2024,wang_OfflineOnline_2022a,jin_Robust_2019}, but are often limited in object diversity and camera viewpoints.
More recent work scales object diversity and sensory richness: Robo360~\cite{liang_Robo360_2023} offers 86 camera views across categories like bags, food, and ropes but lacks high-fidelity annotations and tactile sensing; PokeFlex~\cite{obrist_PokeFlex_2025} introduces volumetric deformable objects and interaction wrenches for poking and dropping actions; HMDO~\cite{xie_HMDO_2023} and DOT~\cite{li_Textureless_2025} target hand manipulation and textureless object tracking, respectively, with detailed surface reconstructions but a restricted range of interaction types.
Simulation-based repositories like ClothesNet~\cite{zhou_ClothesNet_2023} offer massive object diversity but lack real-world physical feedback and interaction complexity. 

In contrast, \shortname provides massive-scale real-world visuotactile sequences, including high-fidelity 2D/3D annotations, 41-view calibrated and synchronized video and tactile data for 198 diverse objects. 
With 23.3 million frames and 215.7 hours of data, it significantly exceeds the scale and sensory richness of existing benchmarks, providing a robust foundation for dynamics evaluation.

\subsection{World Models}
\textbf{Action-conditioned 2D Video Models.} World models enable robots to learn environmental dynamics and plan actions through imagination~\cite{sutton_Dyna_1991, ha_World_2018, lecun_Path_2022, hafner_Dream_2020, hafner_Mastering_2023, wu_DayDreamer_2022}.
Recently, video diffusion models have emerged as a powerful paradigm for world simulation~\cite{blattmann_Stable_2023, nvidia_Cosmos_2025,li_NovaFlow_2025,fu_NovaPlan_2026,du_Video_2023,du_Learning_2023,wan_Wan_2025,kong_HunyuanVideo_2025,team_LongCatVideo_2025,wiedemer_Video_2025,patel_Robotic_2025,cong2025viva,du2026videogpa,agibot-world-contributors_AgiBot_2025}, with action-conditioned variants such as \cite{wang_Interactive_2026}, Vid2World~\cite{huang_Vid2World_2025}, PAN~\cite{team_PAN_2025}, and Cosmos~\cite{nvidia_Cosmos_2025} predicting 3D dynamics \emph{implicitly} in latent space~\cite{lecun_Path_2022,zhou_DINOWM_2025,hafner_Mastering_2023,hansen_TDMPC2_2023,wang2026temporal}.
While these models exhibit remarkable scalability and capture physics from internet-scale pre-training, they often suffer from 3D and temporal inconsistencies during long-horizon predictions~\cite{li_WorldModelBench_2025}.

\textbf{3D World Models.} To incorporate stronger structural priors, 3D dynamics models represent objects using explicit geometry such as meshes or particles~\cite{li_Learning_2019, sanchez-gonzalez_Learning_2020, li_Propagation_2019, zhang_ParticleGrid_2025,huang_PointWorld_2026}, deriving transition functions through either physics-based optimization or data-driven learning.
Physics-based approaches utilize differentiable simulators~\cite{newbury_Review_2024,abou-chakra_RealisSim_2025,chen_EMPM_2026,xian_FluidLab_2022} -- such as DiffCloth~\cite{li_DiffCloth_2022}, PhysTwin~\cite{jiang_PhysTwin_2025a}, and PhysGaussian~\cite{xie_PhysGaussian_2024} -- to embed physical priors like mass conservation, achieving long-term stability but remaining susceptible to state estimation noise.
Conversely, learning-based approaches approximate dynamics using MLPs~\cite{zhang_ParticleGrid_2025}, Graph Neural Networks~\cite{li_Learning_2019, wang_OfflineOnline_2022a, zhang_Dynamic_2024} or Transformers~\cite{huang_ParticleFormer_2025}.
While free from specific simulator inaccuracies~\cite{ai_review_2025}, they typically struggle in low-data scenarios compared to physics-driven methods.

\section{The \shortname Dataset}
\label{sec:dataset}
The \shortname dataset is constructed to address the scarcity of high-fidelity, large-scale real-world data for deformable object modeling.
Unlike existing benchmarks that often rely on synthetic environments or restricted object categories, \shortname provides a dense, synchronized visuotactile record of complex physical interactions across a vast distribution of materials.
198 daily-life objects and 1,980 interaction sequences are included.

\subsection{Visuotactile Capture System}
Each interaction sequence is recorded using a synchronized visuotactile rig designed for $360^{\circ}$ observability~\cite{lu_DiVA360_2023}.
The visual modality consists of RGB videos from $N=41$ cameras, denoted as $\{\mathbf{V}_n\}_{n=1}^N \in \mathbb{R}^{T \times 3 \times H \times W}$, where $T$ represents the number of frames.
The cameras operate at a resolution of $H, W = 720 \times 1280$ at 30 frames per second.
Each camera $n$ is characterized by its intrinsic matrix $\mathbf{K}_n \in \mathbb{R}^{3 \times 3}$ and extrinsic matrix $\mathbf{E}_n \in \mathbb{R}^{4 \times 4}$, representing its fixed pose in the world coordinate system.
This dense spatial coverage is critical for resolving the contact-induced local deformations that are often occluded by end-effectors or the object's own geometry.
Simultaneously, we record tactile signals $\mathbf{T} \in \mathbb{R}^{T \times D}$ from sensors mounted on the bimanual UMI grippers~\cite{chi_Universal_2024,zhu_Touch_2025}.
These signals are software-synchronized with the visual stream at 30~Hz to provide a continuous record of interaction forces and local material response.

\subsection{Object Taxonomy and Diversity}
The dataset features 198 diverse daily-life objects, spanning a wide range of materials and geometric complexities.
This scale -- totaling 1,980 interaction sequences -- represents an order of magnitude increase over existing real-world deformable benchmarks.
To facilitate research into generalizable dynamics, we categorize the object set based on their physical material responses:
\begin{itemize}
\item \textbf{1D Deformables:} 28 ropes, cables, wires, and other linear objects with varying stiffness and thicknesses.
\item \textbf{2D Deformables:} 98 fabrics, cloths, garments, bags, paper-like objects, and other thin-shell materials.
\item \textbf{3D Volumetric Deformables:} 72 plush toys, stuffed animals, foam objects, and squeezable volumetric objects that exhibit significant shape change.
\end{itemize}
More details can be found in the appendix.

\subsection{Interaction Protocol and Scale}
Data collection is performed using tactile-equipped UMI grippers.
The protocol encompasses a wide range of manipulation primitives, including unimanual poking and squeezing, as well as complex bimanual maneuvers such as stretching, folding, and twisting.
For each episode, we record the robot's proprioception (6D pose and openness) alongside $\{\mathbf{V}_n\}$ and $\mathbf{T}$.
This enables precise action reasoning, ensuring the dataset is suitable for training world models that must understand both global object motion and fine-grained contact physics.

\subsection{Dataset Statistics}
\label{subsec:statistics}
The \shortname dataset comprises 198 unique daily-life objects with a total of 1,980 interaction sequences (5 unimanual and 5 bimanual episodes per object).
Our capture setup utilizes 41 camera viewpoints recording at $720p$ resolution and 30 FPS.
This results in 74,850 raw videos \footnote{We filter out videos with unsynchronized or lost frames, causing slight per-viewpoint variation.} and a total of \textbf{23.3 million frames} across all views (614,490 frames per viewpoint on average).
The single-view video duration totals 20,483 seconds, which corresponds to \textbf{215.7 hours} of cumulative multi-view footage.
The average duration of each interaction episode is 10.34 seconds.
Furthermore, synchronized tactile streams are recorded for each interaction episode, providing a rich source of physical contact data.

\section{Annotation Pipeline}
\label{sec:annotation}

\begin{figure*}[t]
  \centering
  \includegraphics[width=\textwidth]{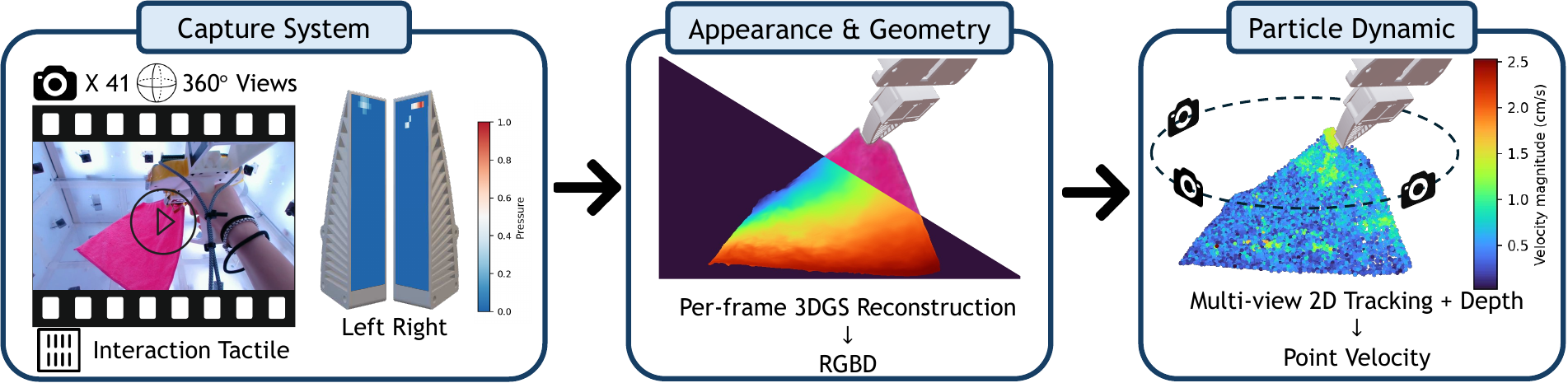}
  \caption{
    \textbf{Overview of our annotation pipeline.}
    We convert multi-view video and tactile streams into dense annotations for deformable object dynamics.
    We first reconstruct high-fidelity dynamic geometry using per-frame 3D Gaussian Splatting. 
    Then, we perform markerless 2D tracking, lift the tracks into 3D, and optimize for temporal and multi-view consistency and physical plausibility.
  }
  \label{fig:pipeline}
\end{figure*}

While Gaussian Splatting-based (GS) approaches~\cite{wu_4D_2024a,luiten_Dynamic_2023} achieve high-fidelity dynamic reconstruction, they are optimized for rendering rather than physical tracking; independent Gaussian motions may result in trajectories that lack temporal consistency~\cite{duisterhof_DeformGS_2024, xie_PhysGaussian_2024}. 
To address this, we decouple per-frame geometry recovery from temporal tracking: GS provides high-quality per-frame geometry, while 2D tracking results from each view are lifted into 3D using this geometry to ensure multi-view consistency.
We further refine the tracking results with tactile signals, which provide ground-truth contact information.
An overview of our pipeline is shown in \Cref{fig:pipeline}.

\subsection{Visuotactile Preprocessing}
The pipeline begins with precise multi-view calibration using ArUco grids, ensuring metric-scale 3D reconstruction. 
Since the raw RGB videos $\{\mathbf{V}_n\}$ are captured with lens distortion, we first undistort all camera streams and refer to the undistorted videos as $\{\mathbf{V}_n\}$ for simplicity throughout the paper.
We track the gripper's 6D pose and openness from multiple views using ArUco markers and aggregate them using the RANSAC algorithm. 
More details regarding the ArUco tracking process can be found in the appendix.

\subsection{Dynamic Geometry Reconstruction}
\label{subsec:reconstruction}
We reconstruct the 3D dynamic geometry of the interacting objects on a per-frame basis using 3D Gaussian Splatting (3DGS)~\cite{kerbl_3D_2023}, which represents a 3D scene as a set of $K$ anisotropic Gaussians.
Each Gaussian $k$ is characterized by its mean position $\boldsymbol{\mu}_k \in \mathbb{R}^3$, a covariance matrix $\boldsymbol{\Sigma}_k = \mathbf{R}_k \mathbf{S}_k \mathbf{S}_k^T \mathbf{R}_k^T$ (decomposed into a rotation matrix $\mathbf{R}_k$ and a scaling matrix $\mathbf{S}_k$ to ensure positive semi-definiteness), an opacity $\alpha_k \in [0, 1]$, and spherical harmonic coefficients encoding view-dependent color $c_k$.
The color and depth maps are obtained through rasterization of the 3D Gaussians~\cite{ye_gsplat_2025}.

To ensure our 3DGS models exclusively capture the object, we deploy segmentation models~\cite{carion_SAM_2025, team_Gemini_2025a} to generate object masks $\mathbf{M}_{\text{obj}, n} \in \{0, 1\}^{T \times H \times W}$ for each camera view $n$.
These masks are then used to filter the RGB videos into segmented, object-only streams $\mathbf{S}_n = \mathbf{V}_n \odot \mathbf{M}_{\text{obj}, n}$, where $\odot$ denotes element-wise multiplication broadcasted over the color channel.
The 3DGS model for each frame is then optimized following the standard objective proposed in \cite{kerbl_3D_2023}, where the loss function $\mathcal{L}_{\text{gs}}$ is defined as a weighted combination of the $\mathcal{L}_1$ loss and the structural similarity index (SSIM) loss: $\mathcal{L}_{\text{gs}} = (1 - \lambda_{\text{gs}}) \mathcal{L}_1 + \lambda_{\text{gs}} \mathcal{L}_{\text{SSIM}}$,
where $\mathcal{L}_1$ is the mean absolute error between the rendered image and the segmented object frame $\mathbf{S}_n$ at time $t$. 
Following the original implementation, we set $\lambda_{\text{gs}} = 0.2$. 
This per-frame reconstruction provides high-fidelity geometry without requiring complex deformation fields. 

\begin{figure*}[t!]
    \centering
    \begin{subfigure}{\textwidth}
        \centering
        \includegraphics[width=\linewidth]{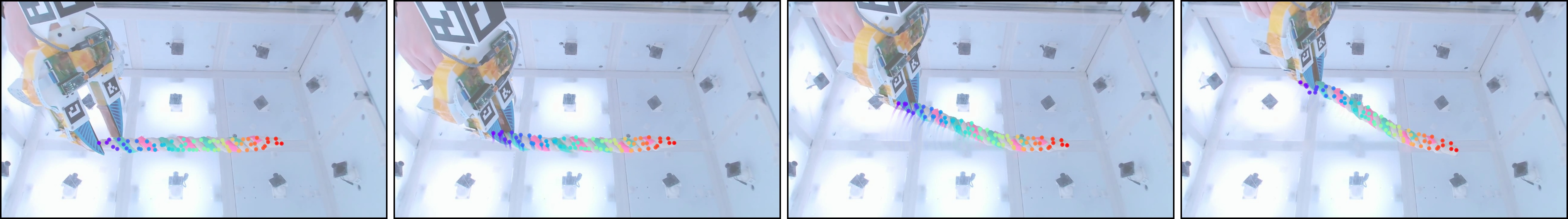}
    \end{subfigure}
    \begin{subfigure}{\textwidth}
        \centering
        \includegraphics[width=\linewidth]{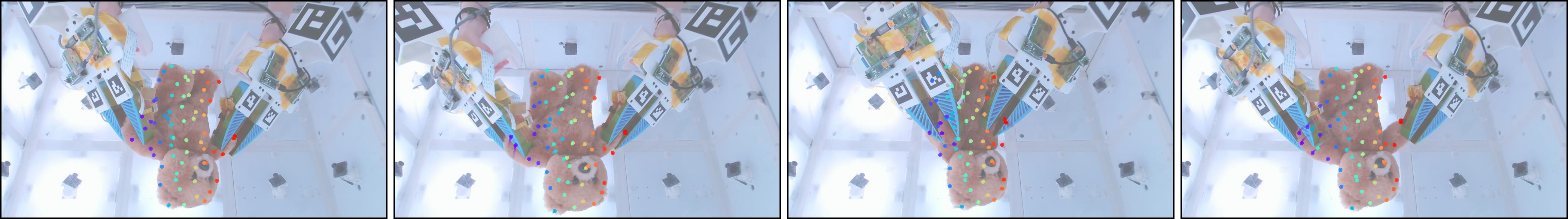}
    \end{subfigure}
    \caption{Visualization of our particle tracking system across different object categories.}
    \label{fig:tracking_qualitative}
\end{figure*}
\subsection{Markerless Particle Tracking}
Once the per-frame geometry is recovered, we perform dense motion estimation using a multi-view 2D-to-3D lifting approach. 
We utilize CoTracker3~\cite{karaev_CoTracker3_2024} to track up to $M=1{,}600$ mask-filtered grid points sampled from the segmented object mask $\mathbf{M}_{\text{obj}, n, t}$ across all camera views, providing persistent 2D trajectories $\mathbf{u}_{n,t} \in \mathbb{R}^{M \times 2}$ for each view $n$ at time $t$.
Following PGND~\cite{zhang_ParticleGrid_2025}, our tracking is performed over 15-frame clips with stride 5 to effectively address invisible points caused by self-occlusions and gripper interactions.
While CoTracker3 offers robust 2D correspondence, 3D lifting is critical to ensure geometric consistency.
For each camera view, we back-project the 2D tracks into 3D space using the rendered depth map $\mathbf{D}_{n,t}$ from our optimized 3DGS. 
The 3D position $\mathbf{P}_{n,t}$ lifted from a 2D track $\mathbf{u}_{n,t}$ is given by:
\begin{equation}
    \mathbf{P}_{n,t} = \mathbf{E}_n^{-1} \mathbf{D}_{n,t}(\mathbf{u}_{n,t}) \mathbf{K}_n^{-1} \tilde{\mathbf{u}}_{n,t},
\end{equation}
where $\tilde{\mathbf{u}}_{n,t}$ is the homogeneous coordinate of $\mathbf{u}_{n,t}$. 
These single-view 3D estimates are often noisy or inconsistent due to occlusions and depth errors. 
To obtain a stable particle trajectory in the global frame, we fuse the multi-view estimates into a unified velocity field $\mathbf{v}_t$ using RANSAC.

However, simple algebraic fusion does not guarantee physical plausibility.
Crucially, we utilize tactile signals to inspect and enforce the physical plausibility of the particle motions during object-gripper interactions.
We minimize a physics-informed objective function $\mathcal{L}_{\text{track}}$ that enforces temporal coherence, local rigidity, spatial smoothness, and tactile consistency. 
First, the shape loss $\mathcal{L}_{\text{shape}}$ is defined as the bidirectional Chamfer distance between the predicted next-step positions and the target 3DGS point cloud $\mathbf{P}^{\text{target}}_{t+1}$, which ensures geometric alignment with the recovered surface. 
To prevent conflicts with contact constraints, particles influenced by tactile sensors $\mathcal{S}_{\text{tactile}}$ are excluded from this loss.
Second, the local rigidity loss $\mathcal{L}_{\text{local}}$ enforces As-Rigid-As-Possible (ARAP) constraints to preserve the local metric structure of the deformable object:
\begin{equation}
    \mathcal{L}_{\text{local}} = \frac{1}{M} \sum_{i} \sum_{j \in \mathcal{N}(i)} (\|\mathbf{P}_{i,t+1} - \mathbf{P}_{j,t+1}\| - l_{ij}^{\text{rest}})^2,
\end{equation}
where $l_{ij}^{\text{rest}}$ is the pre-computed rest length between connected particles $i$ and $j$. 
Third, we apply a Laplacian regularization to encourage spatial smoothness in the velocity field:
\begin{equation}
    \mathcal{L}_{\text{lap}} = \frac{1}{M} \sum_i \left\| \mathbf{v}_{i,t} - \frac{1}{|\mathcal{N}(i)|} \sum_{j \in \mathcal{N}(i)} \mathbf{v}_{j,t} \right\|^2.
\end{equation}
Finally, the tactile loss $\mathcal{L}_{\text{tactile}}$ incorporates reduced-order physics from tactile sensors: 
\begin{equation}
    \mathcal{L}_{\text{tactile}} = \frac{1}{|\mathcal{S}_{\text{tactile}}|} \sum_{i \in \mathcal{S}_{\text{tactile}}} \| \mathbf{v}_{i,t} - \mathbf{v}_{\text{sensor}} \|^2,
\end{equation}
where the contact set $\mathcal{S}_{\text{tactile}}$ is determined by identifying all object particles within a radius $r$ of the activated taxels of the tactile sensors.
Note that $\mathcal{L}_{\text{tactile}}$ assumes localized no-slip only around activated taxels and acts as a soft regularizer rather than a hard constraint.
Since our tactile sensors only measure normal-axis pressure, tangential slip is unobservable. 
Slip events are intentionally avoided during data collection.
The total tracking objective is minimized through gradient descent:
\begin{equation}
    \mathcal{L}_{\text{track}} = \mathcal{L}_{\text{shape}} + \lambda_{\text{local}} \mathcal{L}_{\text{local}} + \lambda_{\text{lap}} \mathcal{L}_{\text{lap}} + \lambda_{\text{tactile}} \mathcal{L}_{\text{tactile}}.
\end{equation}
We set $\lambda_{\text{local}}=20.0$, $\lambda_{\text{lap}}=0.1$, and $\lambda_{\text{tactile}}=1.0$ in all experiments.
This optimization decouples geometry from tracking, ensuring that particles maintain their physical identity even under large deformations and self-occlusions.
We refer to the resulting sequence of configurations, generated by integrating these optimized per-frame velocities starting from the initial particle positions, as the warped point cloud.

\begin{figure*}[t!]
    \centering
    \includegraphics[width=\linewidth]{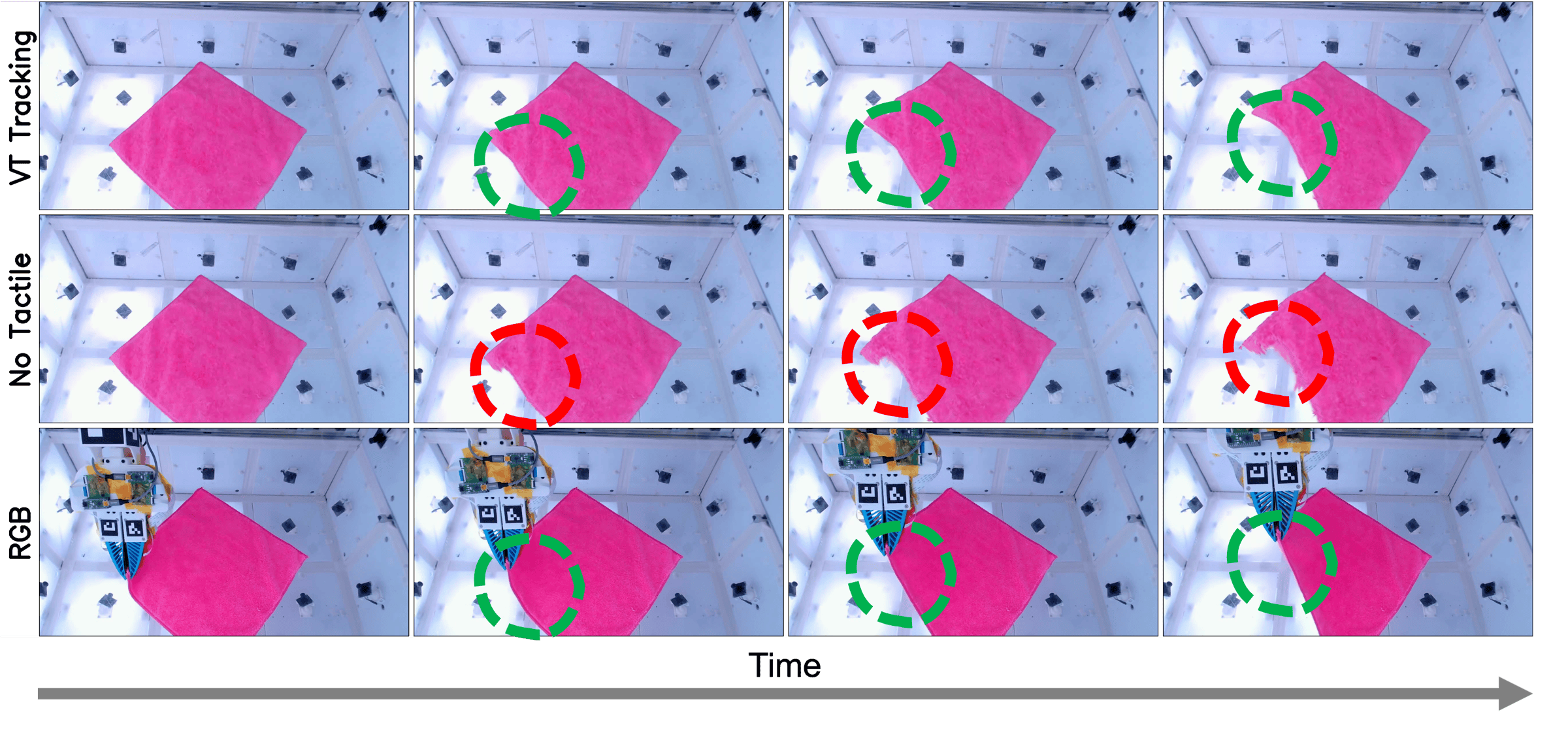}
    \caption{
        \textbf{Benefit of Visuotactile Perception.}
        The three rows compare the warped point clouds, obtained by temporally integrating per-frame tracking results: our full visuotactile approach, tracking without tactile optimization, and the RGB ground truth.
        Our method yields superior tracking accuracy under occlusions caused by the gripper, confirming the advantage of integrating tactile feedback into the tracking pipeline.
    }
    \label{fig:pink_cloth_benefit}
\end{figure*}

\section{Experiments}
\label{sec:experiments}
Our experimental evaluation is designed to demonstrate the utility of \shortname for learning and benchmarking deformable dynamics across multiple scales and modalities.
We first assess the fidelity of our markerless perception pipeline (\cref{sec:recon_experiment}).
Next, we investigate the unique visuotactile coupling in our dataset by predicting contacts from visual observations (\cref{sec:visuotactile_contact}).
Then, we conduct a systematic comparison of state-of-the-art world models~\cite{jiang_PhysTwin_2025a,zhang_ParticleGrid_2025,nvidia_Cosmos_2025,huang_ParticleFormer_2025}, comparing 3D particle dynamics and action-conditioned video models across three levels of generalization: frame, episode, and object (\cref{sec:world_model_benchmarking}).
Finally, we provide a preliminary demonstration of the potential real-world applicability of our dataset by performing robot planning tasks using Model Predictive Control (MPC) on deformable objects (\cref{sec:robot_planning}).
Through these experiments, we aim to provide a comprehensive baseline and reveal critical insights into the future of generalizable deformable world models.

\begin{wraptable}{r}{0.5\textwidth}
    \caption{\small \textbf{Reconstruction Quality.} Performance averaged per category.}
    \centering
    \resizebox{\linewidth}{!}{
        \begin{tabular}{@{} l ccc @{}}
            \toprule
            Category & PSNR $\uparrow$ & SSIM $\uparrow$ & LPIPS $\downarrow$ \\
            \midrule
            1D Deformables & 28.85 & 0.98 & 0.0451 \\
            2D Deformables & 25.77 & 0.95 & 0.0919 \\
            3D Volumetric Deformables & 30.00 & 0.98 & 0.0481 \\
            \textbf{Global Average} & 27.66 & 0.96 & 0.0708 \\
            \bottomrule
        \end{tabular}
    }
    \label{tab:category_gs}
\end{wraptable}

\subsection{Evaluation of Reconstruction and Tracking Quality}
\label{sec:recon_experiment}

We evaluate the rendering performance across the 198 objects in the \shortname dataset (\cref{tab:category_gs}), broken down by object categories (1D, 2D, and 3D volumetric deformables).
We utilize standard image quality metrics, including Peak Signal-to-Noise Ratio (PSNR), Structural Similarity Index (SSIM), and Learned Perceptual Image Patch Similarity (LPIPS), measured on held-out test views.
Our results demonstrate consistently high reconstruction fidelity across all categories, with 3D volumetric objects achieving the highest PSNR of 30.00 dB.
This provides a reliable geometric foundation for the subsequent particle tracking and dynamics modeling tasks.
As shown in \cref{fig:pink_cloth_benefit}, the integration of tactile feedback significantly enhances the tracking accuracy, especially for objects with fully occluded parts during prehensile manipulations.
Quantitatively, the warped point cloud using visuotactile tracking achieves a Chamfer distance error of $2.71 \times 10^{-5} m^2$, five times lower than the $1.41 \times 10^{-4} m^2$ error of the warped point cloud using only visual tracking.
This improvement highlights how the integration of tactile cues provides critical ground-truth signals to maintain tracking accuracy through occlusions.

\subsection{Visuotactile Contact Prediction}
\label{sec:visuotactile_contact}
\begin{wrapfigure}{r}{0.5\textwidth}
    \centering
    \includegraphics[width=\linewidth]{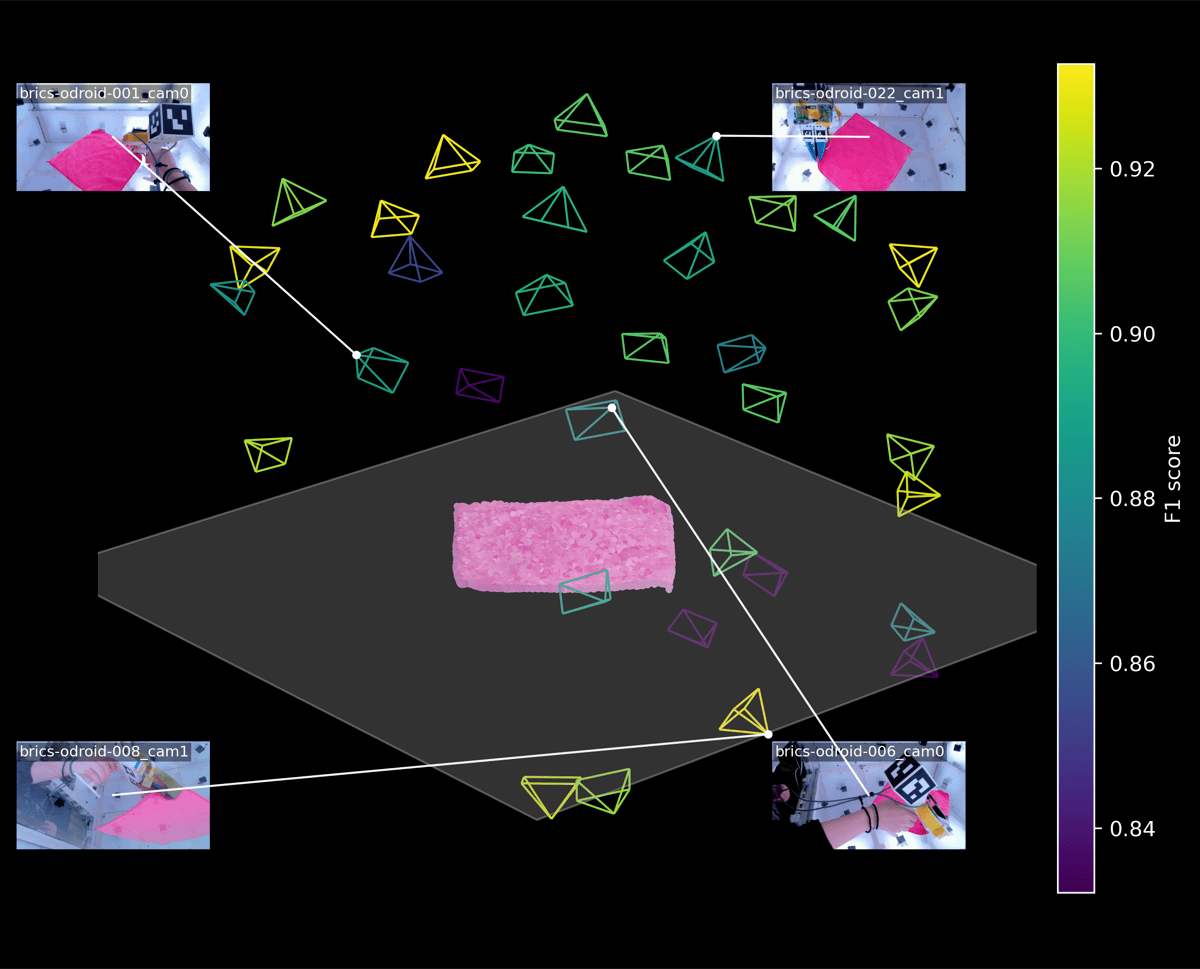}
    \caption{A visualization of the prediction performance from different camera views.}
    \label{fig:contact_prediction}
\end{wrapfigure}

A unique feature of the \shortname dataset is the synchronized tactile modality.
In this experiment, we investigate the feasibility of predicting local tactile contact events solely from visual observations.
We discretize the tactile data into a binary contact/no-contact signal~\cite{li_HyperTaxel_2024,qin_DexPoint_2023,li_ViTaZero_2025,li_VHOP_2025,yin_Rotating_2023}, and train a transformer-based encoder~\cite{oquab_DINOv2_2023,vaswani_Attention_2017} to map the visual stream and robot actions to the expected contact signal.
Our model achieves a mean accuracy of $88.67\%$ across 36 synchronization-filtered views (random guessing $50.31\%$) and an F1-score of $0.8909$ (\cref{fig:contact_prediction}).
The results show that the model reliably captures contact-induced deformations from visual cues, highlighting the dataset's utility in learning the coupling between visual surface changes and internal contact physics.

\subsection{World Model Benchmarking}
\label{sec:world_model_benchmarking}
We benchmark current state-of-the-art world models on our dataset to evaluate their ability to capture complex deformable dynamics.
Our evaluation spans two primary paradigms: predicting 3D dynamics via particle-based models and predicting 2D dynamics via action-conditioned video models.

\textbf{3D World Models.}
We train learning-based 3D world models, such as ParticleFormer~\cite{huang_ParticleFormer_2025} and PGND~\cite{zhang_ParticleGrid_2025}, and optimize the differentiable simulation-based PhysTwin~\cite{jiang_PhysTwin_2025a} using the dense particle trajectories obtained from our annotation pipeline.
We measure the multi-step prediction error using Chamfer distance (CD) and mean squared error (track error) between the ground-truth and the predicted particle positions.
We further render the predicted particle trajectories with first-frame 3DGS via linear blend skinning and calculate the image quality metrics against the ground-truth videos~\cite{jiang_PhysTwin_2025a}.

\textbf{Action-conditioned Video Models.}
We assess the performance of the \emph{pretrained} generative video model Cosmos-Predict 2.5 2B~\cite{nvidia_Cosmos_2025}, which has 2 billion parameters.
As the pretrained Cosmos model does not natively take robot actions as input, we post-train it on the \shortname dataset (according to the evaluation setting) to add robot action conditioning (represented by the 6D wrist pose and gripper openness), enabling it to function as an action-conditioned world model.
Models are evaluated using the same rendering metrics as 3D dynamics models.
More details regarding post-training can be found in the appendix.

While 3D particle models utilize explicit 3D geometry unlike 2D video models, we evaluate them jointly to highlight the trade-offs between structured priors and implicit generation.
This comparison remains practical, as 3D reconstructions are obtainable in real-world robotic deployments~\cite{team_SAM_2025, xiang_Structured_2024,hunyuan3d_Hunyuan3DOmni_2025,lai_Hunyuan3D_2025,xie_CARI4D_2026,wen_BundleSDF_2023}.

\subsubsection{Evaluation settings}

To rigorously verify the generalizability of the evaluated world models, we adopt three distinct evaluation settings: per-episode, multi-episode, and multi-object.

\begin{table*}[t!]
\caption{\small
    \textbf{Quantitative Results on Per-episode Reconstruction \& Resimulation and Future Prediction.}
    We compare the performance of different methods on every episode, divided into reconstruction and resimulation, and future prediction.
    }
\centering
\resizebox{\linewidth}{!}{
\begin{tabular}{@{} l rr rr rr rr rr @{}}
\toprule
Task &  \multicolumn{5}{c}{Reconstruction \& Resimulation} & \multicolumn{5}{c}{Future Prediction} \\
\cmidrule(lr){2-6} \cmidrule(lr){7-11}
Method & CD $\downarrow$ & Track Error $\downarrow$ & PSNR $\uparrow$ & SSIM $\uparrow$ & LPIPS $\downarrow$ & CD $\downarrow$ & Track Error $\downarrow$ & PSNR $\uparrow$ & SSIM $\uparrow$ & LPIPS $\downarrow$ \\
\midrule
PGND & 0.032 & 0.033 & 26.872 & \textbf{0.965} & \textbf{0.046} & 0.073 & 0.073 & 25.296 & 0.963 & 0.050 \\
ParticleFormer & 0.039 & 0.034 & \textbf{26.875} & 0.964 & 0.046 & 0.044 & 0.041 & 26.288 & \textbf{0.964} & \textbf{0.047} \\
PhysTwin & \textbf{0.014} & \textbf{0.021} & 26.783 & 0.963 & 0.048 & \textbf{0.014} & \textbf{0.025} & \textbf{26.574} & \textbf{0.964} & \textbf{0.047} \\
\bottomrule
\end{tabular}
}
\label{tab:frame_generalization}
\end{table*}

\textbf{Per-episode (Frame generalization):} 
Within the same interaction episode, we train the model on the initial $T_{\text{train}}$ frames and evaluate its forecasting capability on the subsequent $T - T_{\text{train}}$ frames to verify its frame generalization.
In \Cref{tab:frame_generalization}, the predicted frames seen during training are referred to as ``reconstruction'' frames, while the predicted frames beyond the training horizon are referred to as ``prediction'' frames.
This setting tests the model's ability to generalize to novel actions given limited dynamic observations.

We observe that the physics-based approach (PhysTwin) outperforms learning-based approaches (PGND and ParticleFormer) in both reconstruction and prediction tasks.
This indicates that explicit physical and structural priors remain critical for accurate dynamics modeling in low-data regimes.
Note that we do not include Cosmos~\cite{nvidia_Cosmos_2025} in this setting as the extremely limited per-episode data is insufficient for the post-training to produce stable predictions.

\begin{table*}[t!]
\caption{\small
    \textbf{Quantitative Results on Multi-episode Generalization.}
    We compare the performance of different methods on reconstruction and future prediction.}
\centering
\resizebox{\linewidth}{!}{
\begin{tabular}{@{} l rr rr rr rr rr @{}}
\toprule
Task &  \multicolumn{5}{c}{Reconstruction \& Resimulation} & \multicolumn{5}{c}{Future Prediction} \\
\cmidrule(lr){2-6} \cmidrule(lr){7-11}
Method & CD $\downarrow$ & Track Error $\downarrow$ & PSNR $\uparrow$ & SSIM $\uparrow$ & LPIPS $\downarrow$ & CD $\downarrow$ & Track Error $\downarrow$ & PSNR $\uparrow$ & SSIM $\uparrow$ & LPIPS $\downarrow$ \\
\midrule
PGND & 0.060 & 0.071 & 25.513 & 0.970 & 0.039 & 0.130 & 0.144 & 23.788 & 0.971 & 0.040 \\
ParticleFormer & \textbf{0.042} & \textbf{0.050} & 26.263 & \textbf{0.971} & 0.038 & \textbf{0.051} & \textbf{0.079} & \textbf{25.203} & \textbf{0.972} & \textbf{0.038} \\
Cosmos & - & - & \textbf{27.748} & 0.964 & \textbf{0.034} & - & - & 24.950 & 0.962 & 0.043 \\
\bottomrule
\end{tabular}
}
\label{tab:multi_episode}
\end{table*}

\textbf{Multi-episode (Episode generalization):} 
As visualized in the figure in the Appendix, for a given object, we train the model on a subset of $E_{\text{train}}$ distinct interaction episodes and test its temporal dynamics prediction on the remaining unseen episodes to verify episode generalization.
This evaluates whether the model has learned the intrinsic physical properties of the object and can generalize to unseen configurations and interaction sequences.
PhysTwin is excluded from this setting as it requires additional per-episode registration to align with different initial configurations, which limits its ability to generalize across episodes without manual intervention.

As seen in \Cref{tab:multi_episode}, we observe a distinct trade-off between 2D video models and 3D particle models.
Cosmos achieves superior reconstruction performance, as training directly in the 2D space allows it to better preserve high-fidelity visual textures and surface details compared to the particle-to-image rendering pipeline.
However, ParticleFormer outperforms in future prediction tasks, which suggests that while the video model excels at visual synthesis, a few interaction episodes are still insufficient for it to fully capture the underlying 3D physical dynamics.
In contrast, the explicit 3D structural priors enable more robust generalization of the object's dynamics to novel interaction sequences.

\begin{figure*}[t!]
    \centering
    \includegraphics[width=\linewidth]{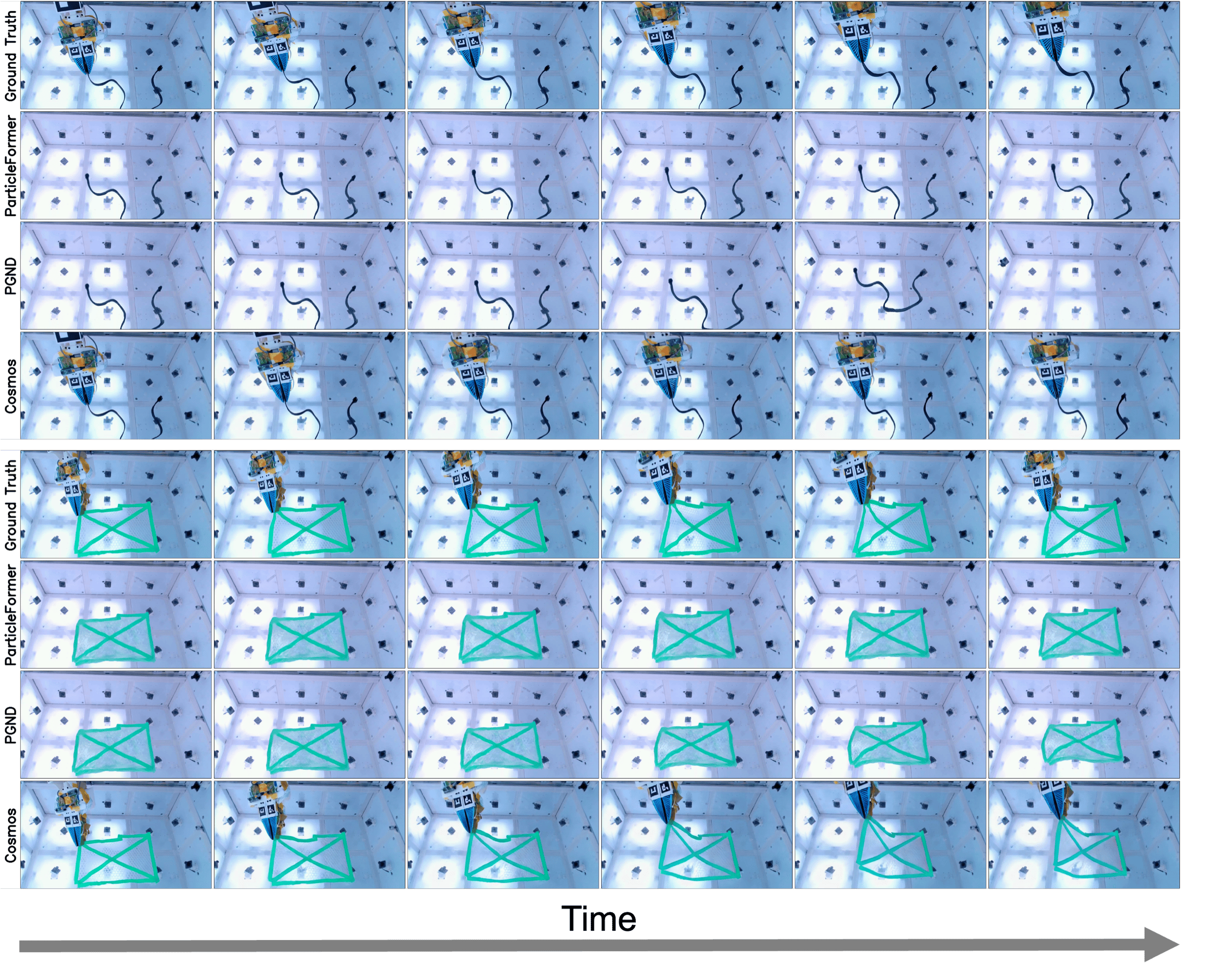}
    \caption{\textbf{Visualization of Multi-Object Generalization}. We show the predicted future frames for the cable (top) and bubble-wrap (bottom) objects, comparing different world models under the zero-shot object generalization setting.}
    \label{fig:object_generalization}
\end{figure*}

\textbf{Multi-object (Object generalization):}
As visualized in \Cref{fig:object_generalization}, we train the model on a set of $O_{\text{train}}$ objects and explicitly test its zero-shot generalization capabilities on the remaining unobserved objects to verify object generalization.
\begin{wraptable}{r}{0.5\textwidth}
    \caption{\small
        \textbf{Quantitative Results on Multi-Object Generalization.}
        We average train/test scores and report them as future prediction.}
    \centering
    \resizebox{0.48\textwidth}{!}{
    \begin{tabular}{@{} l  rrrrr @{}}
    \toprule
    Task &  \multicolumn{5}{c}{Future Pred} \\
    \cmidrule(lr){2-6}
    Method & CD $\downarrow$ & Err $\downarrow$ & PSNR $\uparrow$ & SSIM $\uparrow$ & LPIPS $\downarrow$ \\
    \midrule
    PGND & 0.429 & 0.320 & 22.049 & 0.969 & 0.041 \\
    ParticleFormer & \textbf{0.038} & \textbf{0.048} & 23.312 & \textbf{0.969} & 0.038 \\
    Cosmos & - & - & \textbf{25.042} & 0.958 & \textbf{0.037} \\
    \bottomrule
    \end{tabular}
    }
    \label{tab:multi_object}
\end{wraptable}
This represents the most challenging setting, as it requires the model to infer dynamics and material behaviors for entirely novel object instances based on limited prior knowledge.
Since PhysTwin relies on optimization-based fitting of physical parameters to specific object geometries, it cannot natively generalize to novel object categories.

In this zero-shot setting, Cosmos shows better generalization, outperforming its 3D counterparts in image quality metrics (\Cref{tab:multi_object}), likely because its large-scale pre-training lets it effectively leverage \shortname's visual diversity to infer dynamics for novel object categories.
However, a common failure mode for Cosmos is failing to strictly follow the provided robot commands during long-horizon predictions rather than producing incorrect object dynamics; surprisingly, the resulting dynamics remain physically reasonable most of the time despite this action misalignment, an issue that more fine-tuning data or more sophisticated action representations could help resolve~\cite{team_Evaluating_2026}.
Unlike 2D foundation models, current 3D world models lack massive pre-training, which limits their zero-shot generalization in this setting; while PointWorld~\cite{huang_PointWorld_2026} recently scales up 3D world models, it had not been open-sourced by the time we finished this work.

\subsection{Real-world Robot Planning}
\label{sec:robot_planning}

\begin{figure*}[t!]
    \centering
    \includegraphics[width=\linewidth]{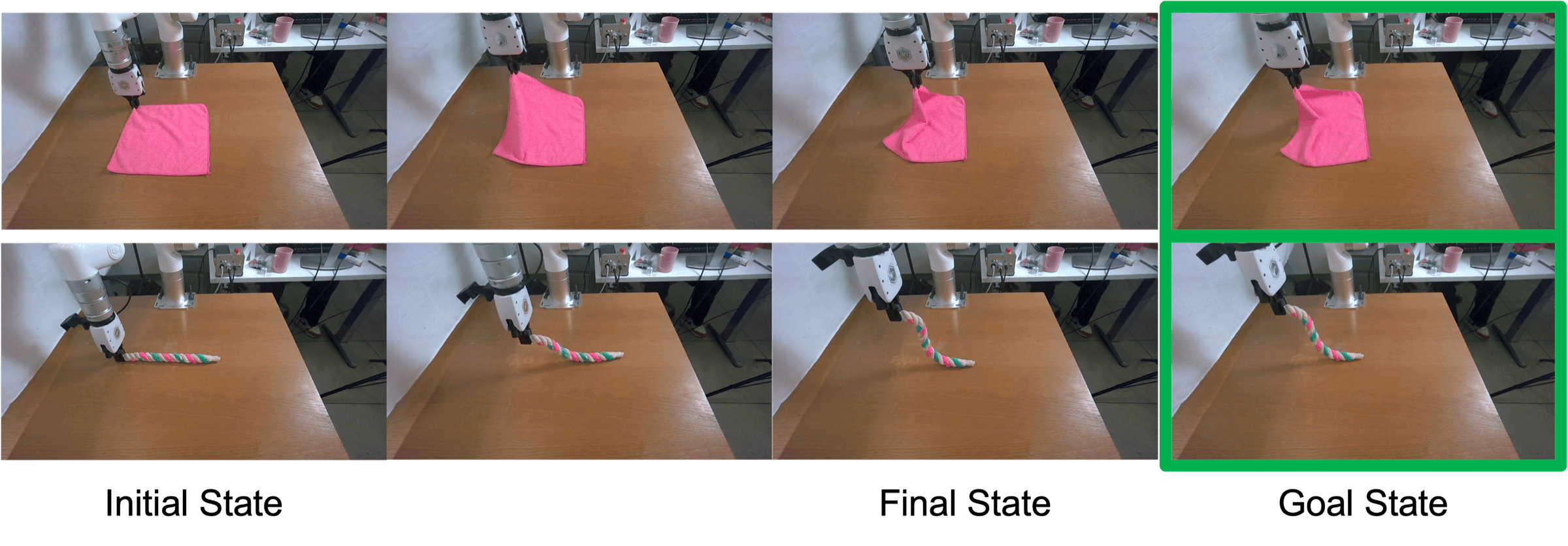}
    \caption{\textbf{Real-world Robot Planning with Model Predictive Control}. We deploy learned models from \shortname in an MPC framework to manipulate various deformable objects toward the goal state.}
    \label{fig:mpc}
\end{figure*}

To evaluate the real-world applicability of our dataset, we demonstrate that object dynamics models learned from \shortname are useful for robotic planning tasks.
Crucially, these experiments are deployed using a completely different robot setup (xArm) located in a different lab environment (\cref{fig:mpc}) in a zero-shot setting.
We utilize representations learned from \shortname via PhysTwin~\cite{jiang_PhysTwin_2025a} within an MPC framework to plan robot actions for achieving a goal state.
We provide preliminary demonstrations indicating the potential real-world applicability, highlighting the utility of \shortname for building generalizable world models with practical robotic applications.

We do not deploy Cosmos for real-world planning due to two primary challenges.
First, video models are more sensitive to appearance differences across environments, and our current post-training scale is insufficient to support such out-of-distribution visual generalization.
Second, designing an effective reward function directly on generated videos remains non-trivial, whereas 3D models can straightforwardly leverage geometric metrics like Chamfer distance~\cite{jiang_PhysTwin_2025a,zhang_ParticleGrid_2025}.

\section{Limitations}
\label{sec:limitations}

\shortname is designed to cover a broad range of everyday deformable interactions, but several difficult cases remain.
Heavy self-occlusion can still reduce tracking quality when large object regions are invisible to most cameras for extended periods.
Highly plastic materials may violate the local rigidity and smoothness assumptions used in our particle optimization, and visible slip at contact can make the tactile no-slip regularizer over-constrain nearby particles.
Because our tactile sensors measure normal-axis pressure, they cannot directly identify micro-slip, so future extensions should incorporate richer tactile sensing and expand the object set toward more extreme unstructured materials.

\section{Conclusion}
\label{sec:conclusion}

We presented \shortname, a large-scale multi-view visuotactile dataset for modeling real-world deformable object dynamics.
By developing a novel markerless 3D tracking pipeline, we extracted high-fidelity particle trajectories and dynamic geometry representations across a diverse set of 198 daily-life objects.
Leveraging this data, we systematically benchmarked state-of-the-art 3D particle dynamics models against 2D action-conditioned video generation models.
Our evaluations revealed a critical trade-off: while 3D particle models benefited from robust structural priors in low-data regimes, video models leveraged massive pre-training for stronger zero-shot visual generalizability.
Designing effective reward functions for video models also proved non-trivial due to their lack of explicit 3D geometry, complicating their direct use in model-based planning.
Despite these challenges, we provide a preliminary demonstration of our dataset's utility by deploying 3D particle models in a real-world Model Predictive Control setup for deformable object manipulation.
Ultimately, \shortname provides a foundational benchmark that we hope will inspire future research toward more generalizable, physically grounded, and scalable world models for robotics.

\section*{Acknowledgements}
This work was supported by the Office of Naval Research (ONR) under REPRISM MURI N000142412603, ONR grant N00014-22-1-2592, and ONR DURIP grant N00014-23-1-2804.
This work was also partially supported by NSF Award \#2409661, Samsung Research America, an Amazon Research Award (Fall 2024), and the Robotics and AI Institute.
Hongyu Li was partially funded by Brown University's Research Mobility Fellowship.
Wanjia Fu was partially funded by Brown University's Randy Pausch Undergraduate Research Fellowship.
This article solely reflects the opinions and conclusions of its authors and should not be interpreted as necessarily representing the official policies, either expressed or implied, of the sponsors.

\bibliographystyle{splncs04}
\bibliography{hongyu_zotero, custom}

\newpage
\appendix

\section{Gripper Pose and Openness Tracking}

\label{sec:aruco_tracking}

To accurately track the gripper's 6D pose and openness during interaction, we attach a set of ArUco markers (from the \texttt{DICT\_4X4\_250} dictionary) to each gripper. 
For each gripper, four markers are placed on the wrist/base (IDs 2--5 or 10--13) and four markers are placed on the fingers (two per finger; IDs 0, 6 and 1, 7 or 8, 14 and 9, 15). 
Each marker $i$ has a 6D offset $\mathbf{T}_{m \to g}^{(i)}$ relative to the gripper's coordinate frame, measured in the CAD software.

For each frame $t$, the tracking pipeline consists of the following steps:
\begin{enumerate}
    \item \textbf{Per-view Detection}: Each camera $c$ detects visible markers and estimates their 6D pose $\mathbf{T}_{c \to m}^{(i, c)}$ in the camera frame using \texttt{solvePnP} with sub-pixel corner refinement. 
    These poses are transformed into the world frame using the calibrated camera extrinsics $\mathbf{T}_{w \to c}$:
    \begin{equation}
        \mathbf{T}_{w \to g}^{(i, c)} = \mathbf{T}_{w \to c} \cdot \mathbf{T}_{c \to m}^{(i, c)} \cdot \mathbf{T}_{m \to g}^{(i)}
    \end{equation}
    \item \textbf{Robust Multi-view Fusion}: We aggregate all estimates $\mathbf{T}_{w \to g}^{(i, c)}$ from all cameras and markers. 
    A RANSAC-based filtering process is employed to identify and remove outlier detections caused by partial occlusions or detection noise. 
    The inlier poses are then averaged to obtain a robust estimate for the gripper wrist pose $\mathbf{T}_{w \to g}^{\text{wrist}}$ and finger poses $\mathbf{T}_{w \to g}^{\text{left}}$ and $\mathbf{T}_{w \to g}^{\text{right}}$. 
    Translation is averaged linearly, while rotation is averaged by computing the mean of the rotation matrices via SVD to ensure the resulting matrix remains in $SO(3)$.
    \item \textbf{Gripper Openness}: The gripper openness $o_t$ is computed as the Euclidean distance between the tracked finger positions:
    \begin{equation}
        o_t = \| \mathbf{t}_{w \to g}^{\text{left}} - \mathbf{t}_{w \to g}^{\text{right}} \|_2
    \end{equation}
\end{enumerate}

\section{Preliminaries}
\label{sec:preliminaries}
We provide additional details on the deformable object-centric world models we use in this work.
Modeling deformable dynamics can be approached through two primary paradigms: predicting 2D dynamics via video models and predicting explicit 3D dynamics via particle models.
2D dynamics models, or video world models, learn to predict future observations directly in the image or latent space.
In contrast, 3D dynamics models represent objects using structured geometric entities, most commonly particles, to capture their physical state and dynamics.
These models further differ in their transition mechanisms, ranging from data-driven learning-based neural architectures to optimization-based differentiable physical simulators.

\subsection{Action-conditioned Video Models}
Action-conditioned video world models aim to predict future video frames $\mathcal{F}_{t+1:t+k}$ based on past observations $\mathcal{F}_{1:t}$ and robot actions $\mathcal{A}_{1:t+k}$.
Modern architectures typically follow the Latent Diffusion Model (LDM)~\cite{blattmann_Stable_2023} paradigm, which operates in a compressed latent space $\mathbf{z}$ to improve computational efficiency.
A Variational Autoencoder (VAE)~\cite{wu_Improved_2025} first encodes an RGB video $\mathbf{x} \in \mathbb{R}^{T \times 3 \times H \times W}$ into a lower-dimensional latent representation $\mathbf{z} = \mathcal{E}(\mathbf{x})$ and reconstructs it via $\hat{\mathbf{x}} = \mathcal{D}(\mathbf{z})$.
The dynamics are then modeled by a Diffusion Transformer (DiT)~\cite{peebles_Scalable_2023}, which learns to reverse a Gaussian denoising process in the latent space.
The model predicts the noise $\epsilon_\theta$ added to the latent state at timestep $\tau$ of the diffusion process:
\begin{equation}
    \mathcal{L} = \mathbb{E}_{\mathbf{z}, \epsilon, \tau, \mathcal{A}} \left[ \| \epsilon - \epsilon_\theta(\mathbf{z}_\tau, \tau, \mathbf{z}_{1:t}, \mathcal{A}) \|^2 \right],
\end{equation}
where $\mathbf{z}_\tau$ is the noisy latent and the diffusion is conditioned on past latents and future actions.
By learning from internet-scale data, these models capture complex visual and physical priors.

Specifically, in our evaluation, we utilize the Cosmos Predict 2.5 model~\cite{nvidia_Cosmos_2025}.
Cosmos Predict 2.5 is a state-of-the-art DiT-based video model that achieves high-fidelity future frame generation.
To equip the pre-trained Cosmos Predict 2.5 with action conditioning, we post-train the model on our dataset.
We inject the robot's 7D action sequence (6D wrist pose and gripper openness) into the DiT blocks via cross-attention.
This process enables the model to accurately simulate the future states of deformable objects conditioned on specific robot interventions.

\subsection{Particle-based Dynamics Models}
Particle-based dynamics models represent a deformable object as a set of $N$ particles $\mathcal{P} = \{(\mathbf{p}_i, \mathbf{v}_i, \mathbf{a}_i)\}_{i=1}^N$, where $\mathbf{p}_i, \mathbf{v}_i \in \mathbb{R}^3$ are the position and velocity, and $\mathbf{a}_i$ represents particle-specific attributes such as mass or material properties.
The system state $\mathcal{G}_t$ evolves according to a transition function $\mathcal{T}$ such that $\mathcal{G}_{t+1} = \mathcal{T}(\mathcal{G}_t, \mathcal{A}_t)$.

\textbf{Learning-based Dynamics.}
Learning-based approaches approximate $\mathcal{T}$ using neural networks trained on large-scale trajectory data.
Graph Neural Networks (GNNs)~\cite{sanchez-gonzalez_Learning_2020, li_Learning_2019} model particles as nodes in a graph and represent physical interactions (e.g., collisions, internal stress) as message-passing over edges.
In our evaluation, we utilize PGND (Particle-Grid Neural Dynamics)~\cite{zhang_ParticleGrid_2025}, which proposes a hybrid framework combining particles for tracking object shape with spatial grids to ensure spatial continuity and computational efficiency. PGND can learn the dynamics of deformable objects directly from real-world videos and employs 3D Gaussian Splatting for dynamic visual rendering.
More recently, the Transformer-based model ParticleFormer~\cite{huang_ParticleFormer_2025} has been employed to capture long-range dependencies and complex multi-object interactions by treating the set of particles as a sequence of tokens.
It leverages multi-head self-attention to infer interactions without relying on predefined adjacency graphs, and is trained using a hybrid point cloud reconstruction loss (combining Chamfer and Hausdorff distances) directly on sensor streams.
These models learn a displacement function $\Delta \mathbf{p} = f_\phi(\mathcal{G}_t, \mathcal{A}_t)$, updating states via $\mathbf{p}_{t+1} = \mathbf{p}_t + \Delta \mathbf{p}$.

\textbf{Physics-based Differentiable Simulation.}
Physics-based models enforce physical laws by embedding a differentiable simulator into the learning loop.
A common representation for deformable objects is the spring-mass system~\cite{jiang_PhysTwin_2025a, zhong_Reconstruction_2025}, where particles are connected by elastic springs.
The force $\mathbf{F}_i$ on particle $i$ is calculated as the sum of internal spring forces, damping, and external forces:
\begin{equation}
    \mathbf{F}_i = \sum_{j \in \mathcal{N}(i)} k_{ij} (\|\mathbf{p}_j - \mathbf{p}_i\| - l_{ij}) \frac{\mathbf{p}_j - \mathbf{p}_i}{\|\mathbf{p}_j - \mathbf{p}_i\|} - \gamma(\mathbf{v}_i - \mathbf{v}_j) + \mathbf{F}_{ext},
\end{equation}
where $k_{ij}$ is the spring stiffness, $l_{ij}$ is the rest length, and $\gamma$ is the damping coefficient.
The state is updated via numerical integration (e.g., semi-implicit Euler): $\mathbf{v}_{t+1} = \mathbf{v}_t + \Delta t \mathbf{F}/m$ and $\mathbf{p}_{t+1} = \mathbf{p}_t + \Delta t \mathbf{v}_{t+1}$.
Differentiable simulation allows for the optimization of physical parameters by backpropagating gradients from the prediction error back to the initial state and material properties.
In our evaluation, we benchmark PhysTwin~\cite{jiang_PhysTwin_2025a}, which builds an interactive physical digital twin from sparse videos. 
It integrates a spring-mass representation for physical simulation with generative shape models for full geometry recovery and 3D Gaussian Splats for rendering. 
To handle partial observations and large parameter spaces, they utilize a hierarchical sparse-to-dense optimization strategy, enabling the accurate estimation of both discrete topological structures and continuous physical properties.

In our benchmarking experiments, we utilize the official open-source implementations for Cosmos Predict 2.5, PhysTwin, and PGND. 
As ParticleFormer was not open-sourced at the time of this project, we faithfully reproduced their method following the implementation details provided in their paper.

\begin{table}[t!]
\renewcommand{\arraystretch}{1.0}
\setlength{\tabcolsep}{3pt}

\centering
\resizebox{\textwidth}{!}{%
\begin{tabular}{lcccccrrrp{2.5cm}p{3.5cm}}
\toprule
\textbf{Dataset}  & \textbf{Mesh}  & \textbf{Calib.} & \textbf{Markerless} & \textbf{Tactile} &  \textbf{360$^{\circ}$}  &  \textbf{Views} & \textbf{\# obj.} & \textbf{\# frames} & \textbf{Action} & \textbf{Deformable Categories} \\
\midrule
ClothSim2Real~\cite{blanco-mulero_Benchmarking_2024} & \cmark & \xmark & \cmark & \xmark & \xmark & 1 & 3 & $\sim$4.9k & drag, fling & cloth \\
ClothTrack~\cite{coltraro_Tracking_2025} & \xmark & \xmark & \xmark & \xmark & \xmark & 5 & 8 & $\sim$173k & collide, hit, shake, twist & cloth \\
DDER~\cite{chen_Differentiable_2024}  & \xmark  & \xmark & \xmark & \xmark & \xmark & 11 & 5 & $\sim$1.8M & bend, drag, lift, shake, twist  & cable, rope  \\
Robo360~\cite{liang_Robo360_2023}  & \xmark & \cmark & \cmark & \xmark & \cmark & 86 & 17 & $\sim$2M & break, drop, fold, lift, push, tear, tie & bag, ball, bread, cable, cloth, food, rope \\
PokeFlex~\cite{obrist_PokeFlex_2025} & \cmark & \cmark & \cmark & \cmark & \cmark & 6 & 18 & 21.3k & drop, poke & foam, paper roll, pillow, plush toys, sponge, 3D printed objects \\
DOT~\cite{li_Textureless_2025} & \cmark & \cmark & \xmark & \xmark & \cmark & 42 & 22 & $\sim$689k & bend, blow, drag, fold, lift, poke, stretch, tilt, twist & cloth, paper, rope \\
HMDO~\cite{xie_HMDO_2023} & \cmark & \cmark & \cmark & \xmark & \cmark & 10 & 12 & 21.6k & pinch, press & plush toys\\
HCOS~\cite{garcia-camacho_Household_2022} & \cmark & \xmark & \cmark & \xmark & \xmark & 1 & 27 & 64 & flattened, folded, grasped  & cloth \\
PhysTwin~\cite{jiang_PhysTwin_2025a} & \xmark & \cmark & \cmark & \xmark & \xmark & 3 & 11 & 7.2k & lift, stretch, press, push & cloth, package, plush toys, rope \\
PGND~\cite{zhang_ParticleGrid_2025} & \cmark & \cmark & \cmark & \xmark & \cmark & 4 & 8 & $\sim$1M & close, drag, lift, open, stretch, tear & box, bread, cloth, paperbag, plush toys, rope \\
\shortname (Ours) & \cmark & \cmark & \cmark & \cmark & \cmark & 41 & 198 & $\sim$23.3M & bend, close, drag, fold, lift, open, poke, press, roll, squeeze, stretch, twist, wave & bag, ball, box, bread, cable, chain, clay, cloth, dice, foam, garments, paper, pillow, plush toys, rope, sponge, thread \\
\bottomrule
\end{tabular}%
}
\caption{
  \textbf{Comparison of real-world datasets containing deformable objects.} 
  We benchmark existing datasets based on key modalities for learning dynamics, including availability of 3D mesh data (Mesh), camera calibration (Cam Calib), \emph{markerless} (whether tracking does NOT rely on physical markers), inclusion of tactile data (Tactile), multi-view coverage (360$^{\circ}$ View), and number of camera views (Views). 
  We also report the scale of each dataset in terms of unique objects (\# of objects) and total frames (\# of frames), along with the manipulation actions and object categories included. 
  `\cmark' means containing a specific modality, and `\xmark' means the opposite.
}
\label{table:dataset_full}
\end{table}

\begin{figure*}[t!]
    \centering
    \begin{minipage}[c]{0.02\linewidth}
        \centering
        \rotatebox{90}{Rope}
    \end{minipage}
    \hfill
    \begin{minipage}[c]{0.96\linewidth}
        \centering

        \begin{subfigure}{0.24\linewidth}
            \centering
            \includegraphics[width=\linewidth]{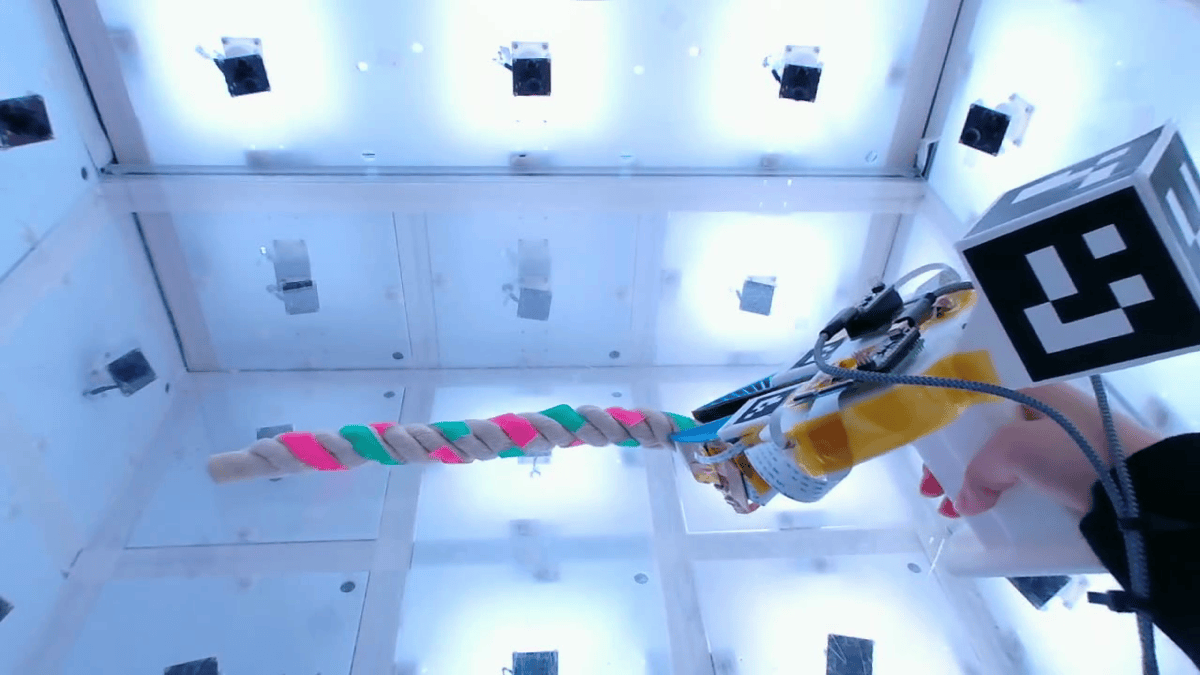}
        \end{subfigure}
        \hfill
        \begin{subfigure}{0.24\linewidth}
            \centering
            \includegraphics[width=\linewidth]{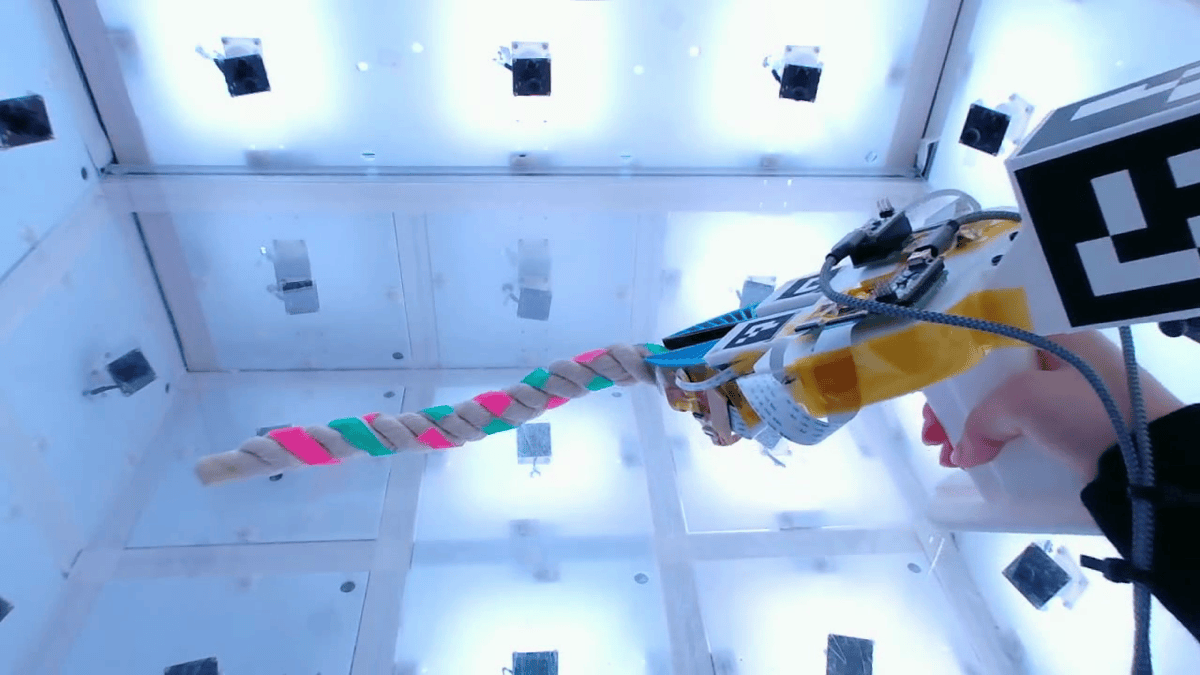}
        \end{subfigure}
        \hfill
        \begin{subfigure}{0.24\linewidth}
            \centering
            \includegraphics[width=\linewidth]{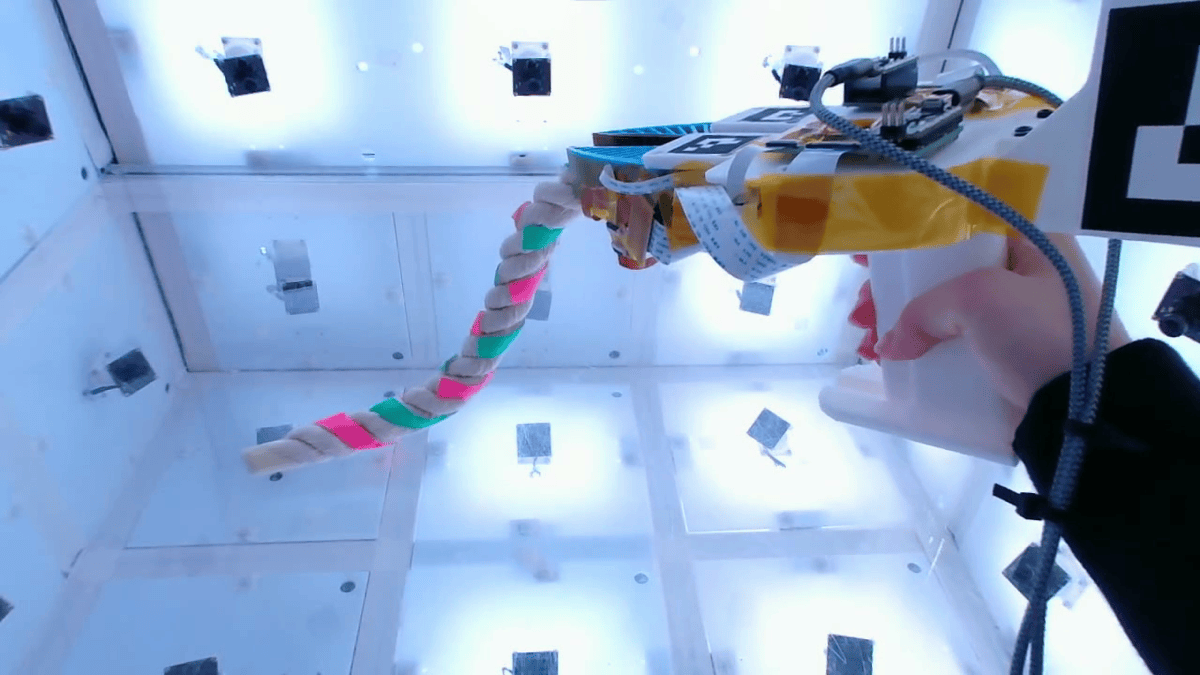}
        \end{subfigure}
        \hfill
        \begin{subfigure}{0.24\linewidth}
            \centering
            \includegraphics[width=\linewidth]{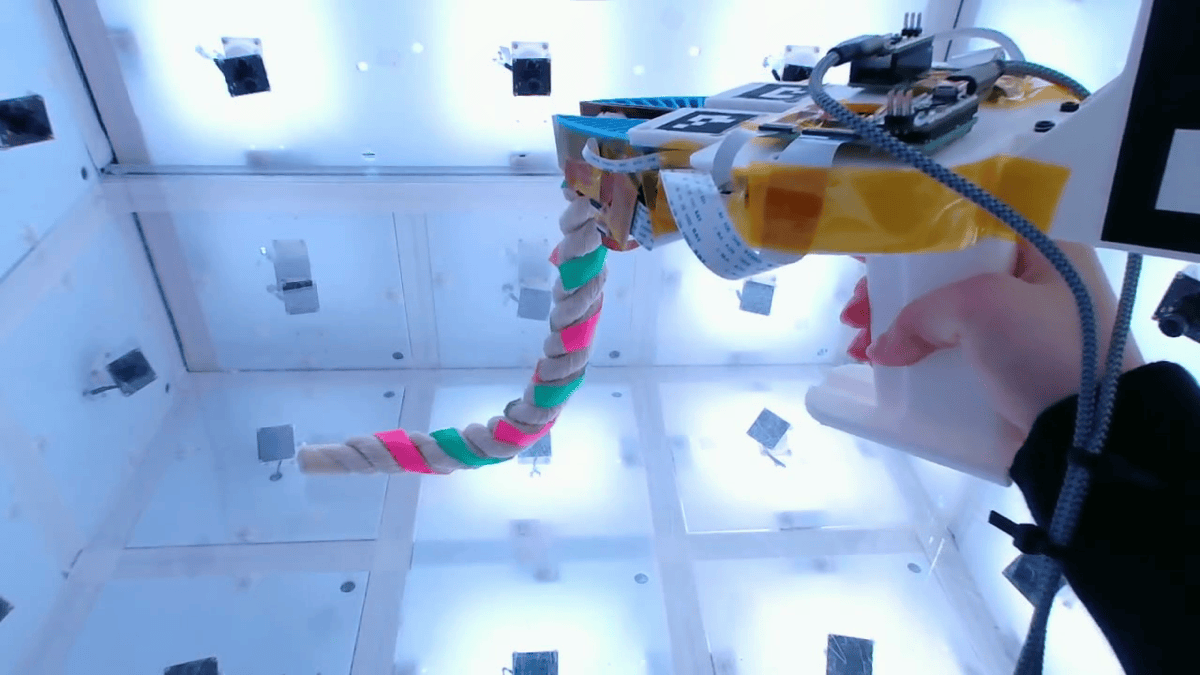}
        \end{subfigure}

        \vspace{0.5em}

        \begin{subfigure}{0.24\linewidth}
            \centering
            \includegraphics[width=\linewidth]{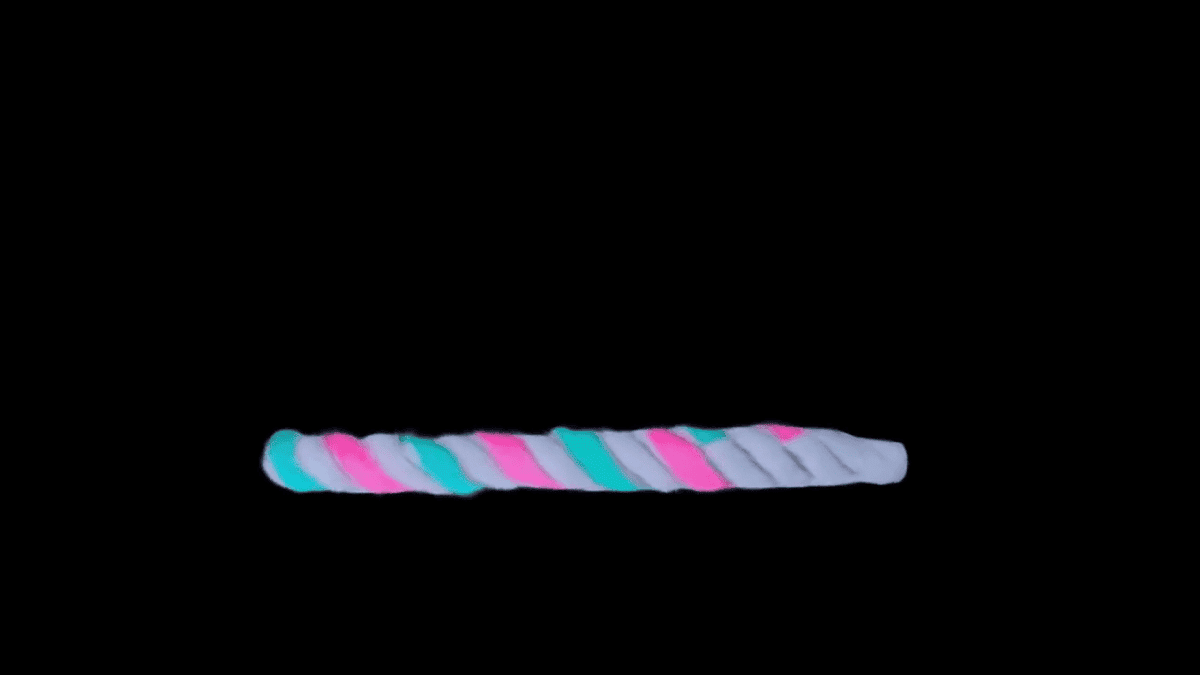}
        \end{subfigure}
        \hfill
        \begin{subfigure}{0.24\linewidth}
            \centering
            \includegraphics[width=\linewidth]{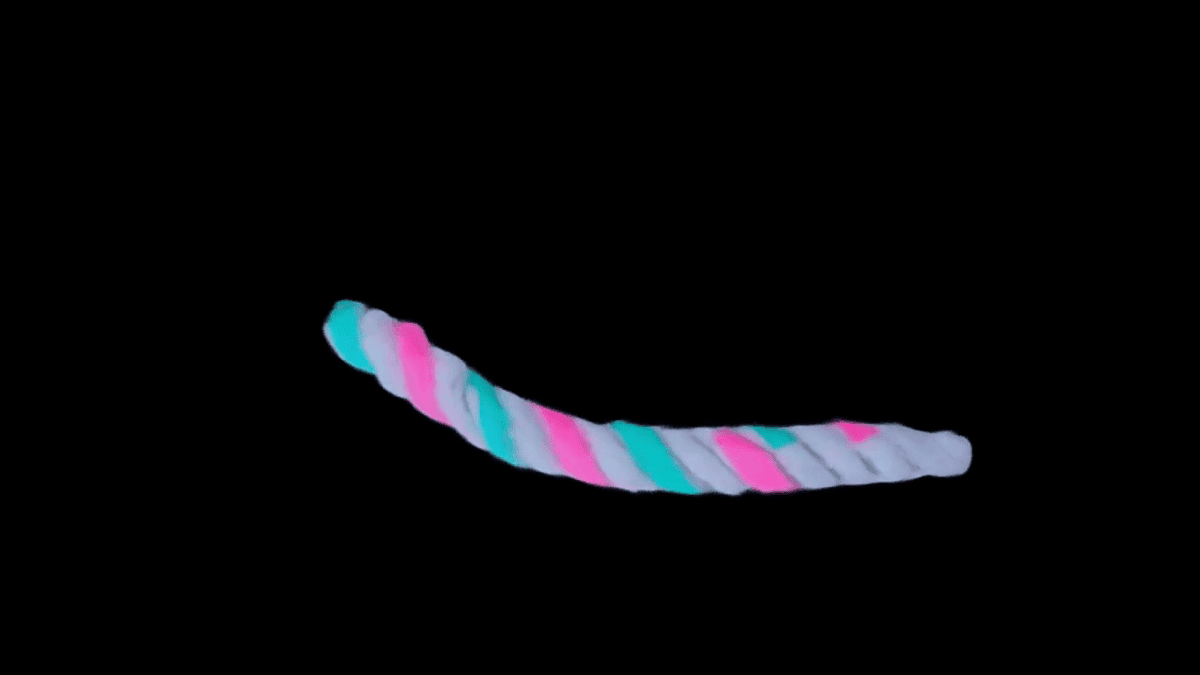}
        \end{subfigure}
        \hfill
        \begin{subfigure}{0.24\linewidth}
            \centering
            \includegraphics[width=\linewidth]{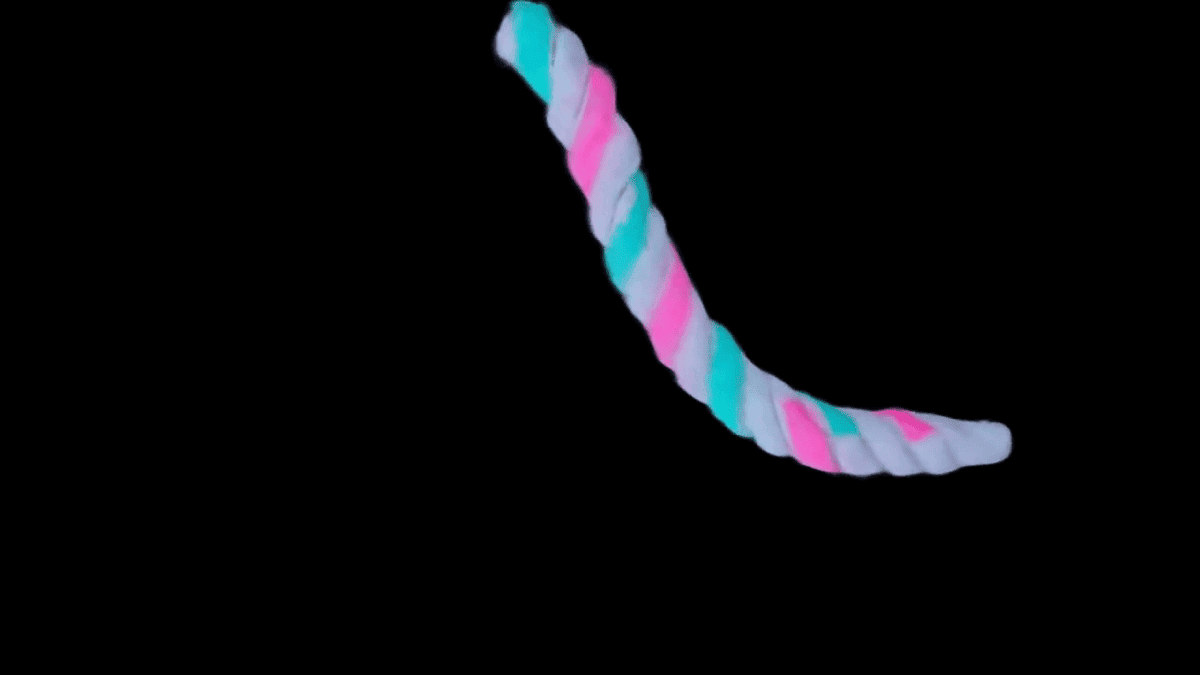}
        \end{subfigure}
        \hfill
        \begin{subfigure}{0.24\linewidth}
            \centering
            \includegraphics[width=\linewidth]{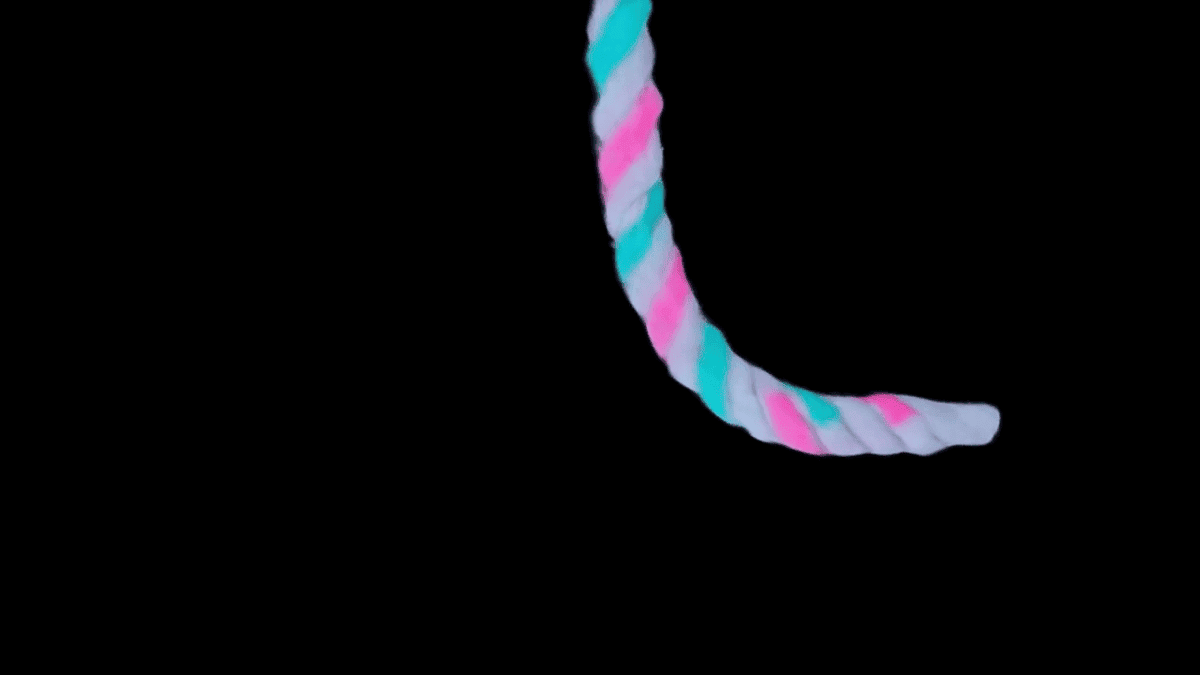}
        \end{subfigure}
    \end{minipage}

    \vspace{0.8em}

    \begin{minipage}[c]{0.02\linewidth}
        \centering
        \rotatebox{90}{Cloth}
    \end{minipage}
    \hfill
    \begin{minipage}[c]{0.96\linewidth}
        
        \begin{subfigure}{0.24\linewidth}
            \centering
            \includegraphics[width=\linewidth]{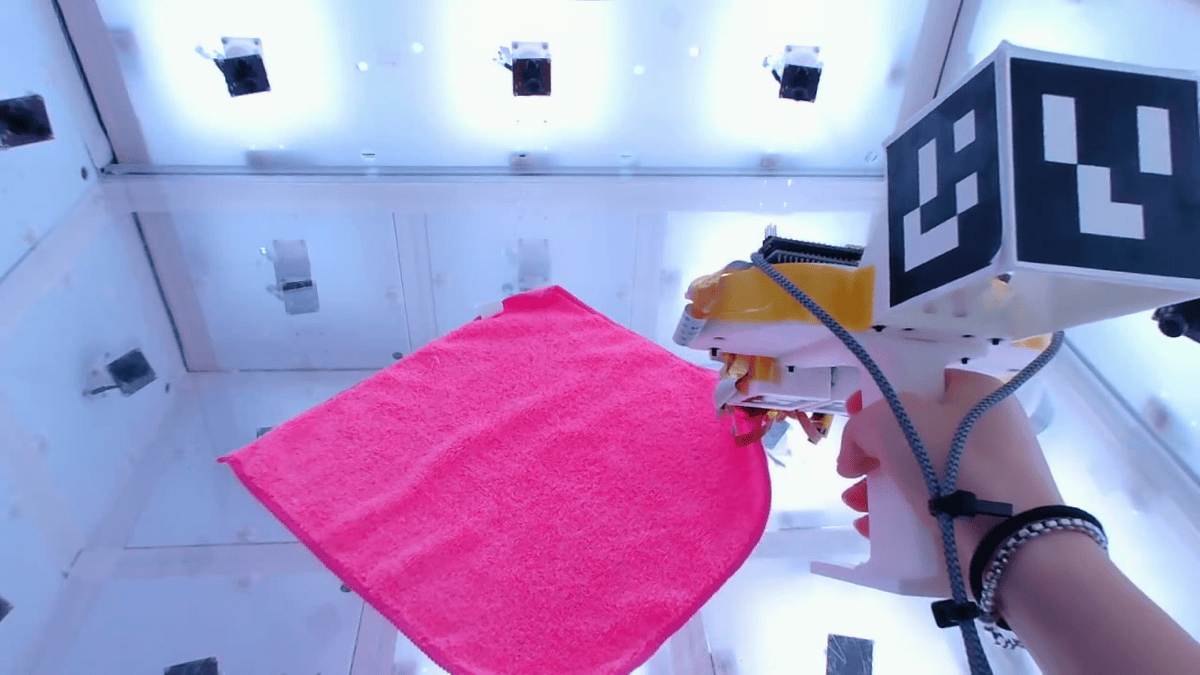}
        \end{subfigure}
        \hfill
        \begin{subfigure}{0.24\linewidth}
            \centering
            \includegraphics[width=\linewidth]{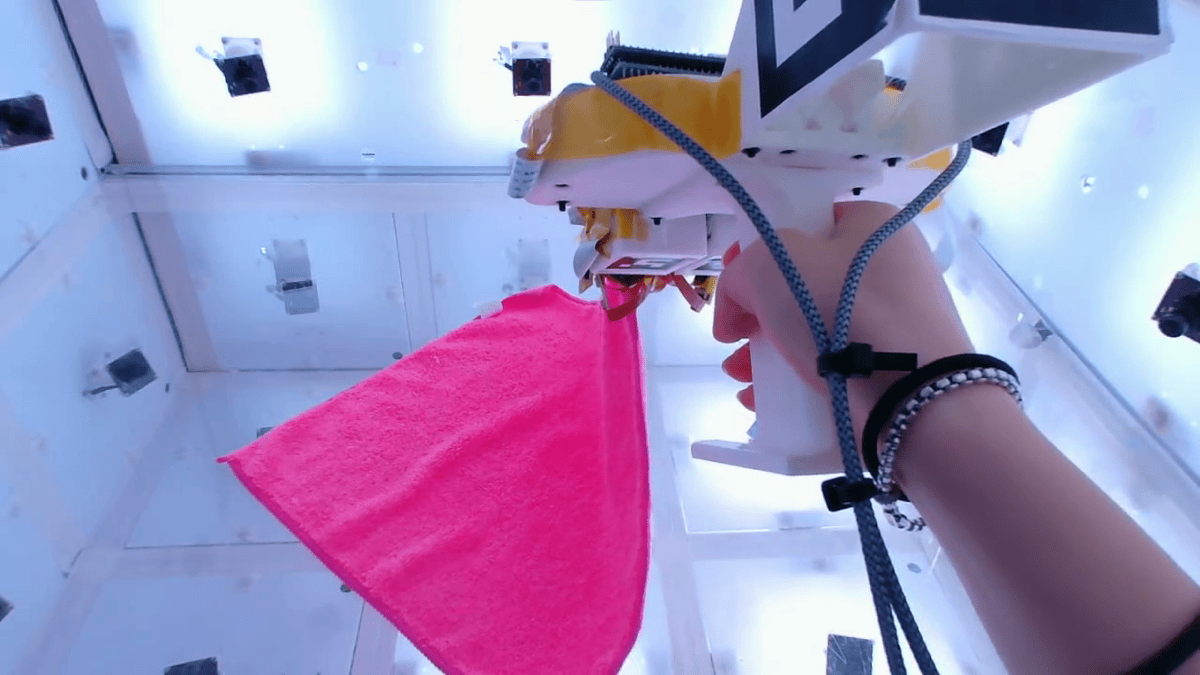}
        \end{subfigure}
        \hfill
        \begin{subfigure}{0.24\linewidth}
            \centering
            \includegraphics[width=\linewidth]{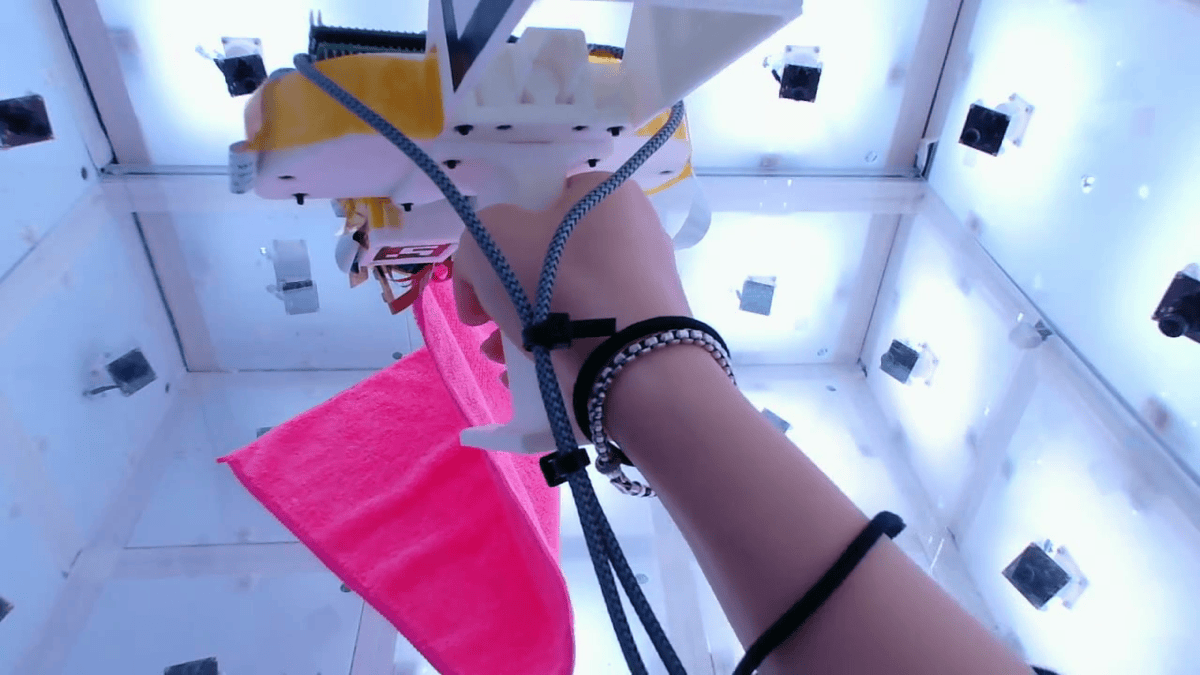}
        \end{subfigure}
        \hfill
        \begin{subfigure}{0.24\linewidth}
            \centering
            \includegraphics[width=\linewidth]{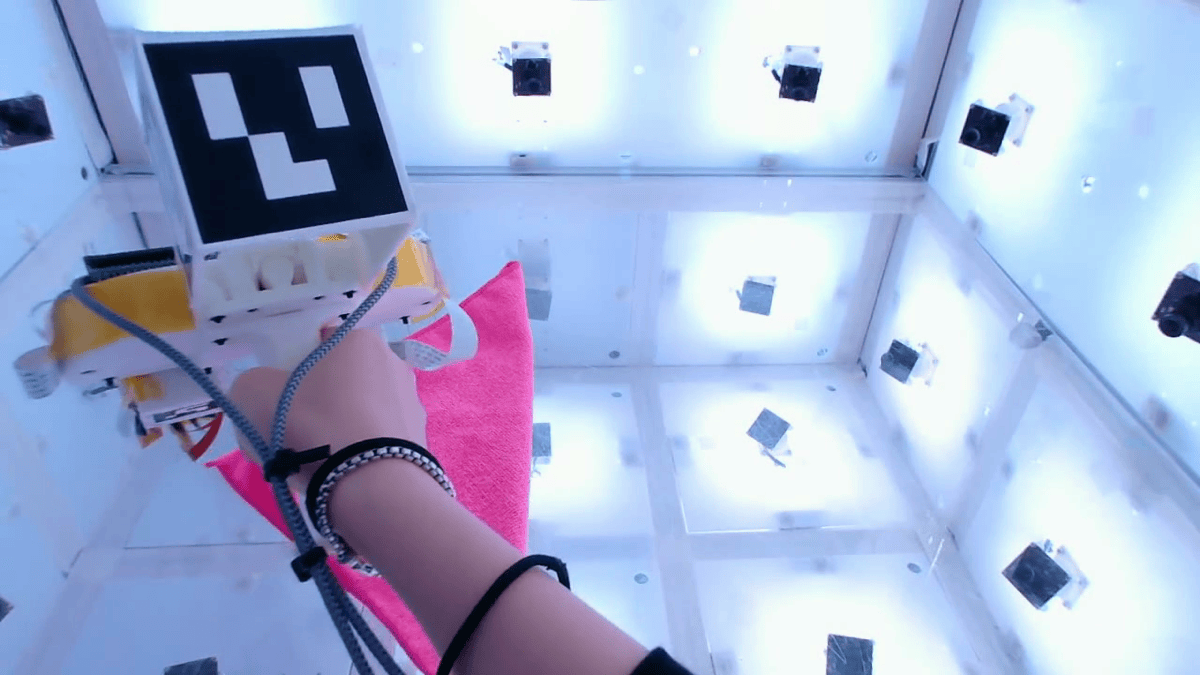}
        \end{subfigure}

        \centering

        \vspace{0.5em}

        \begin{subfigure}{0.24\linewidth}
            \centering
            \includegraphics[width=\linewidth]{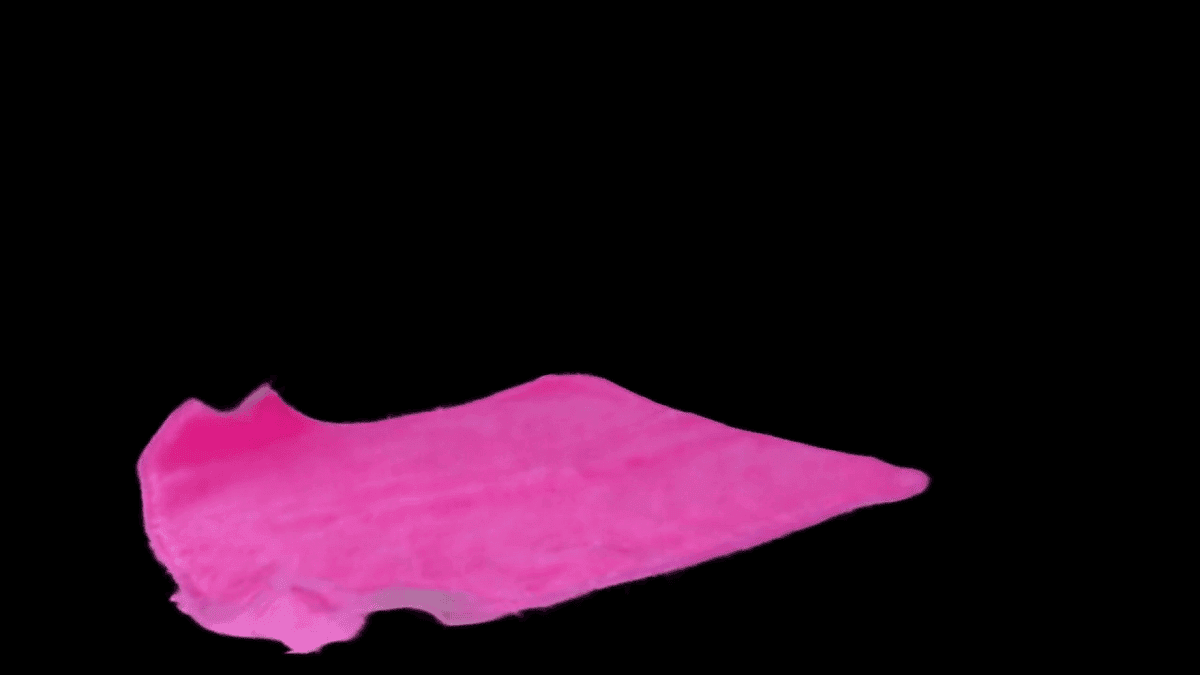}
        \end{subfigure}
        \hfill
        \begin{subfigure}{0.24\linewidth}
            \centering
            \includegraphics[width=\linewidth]{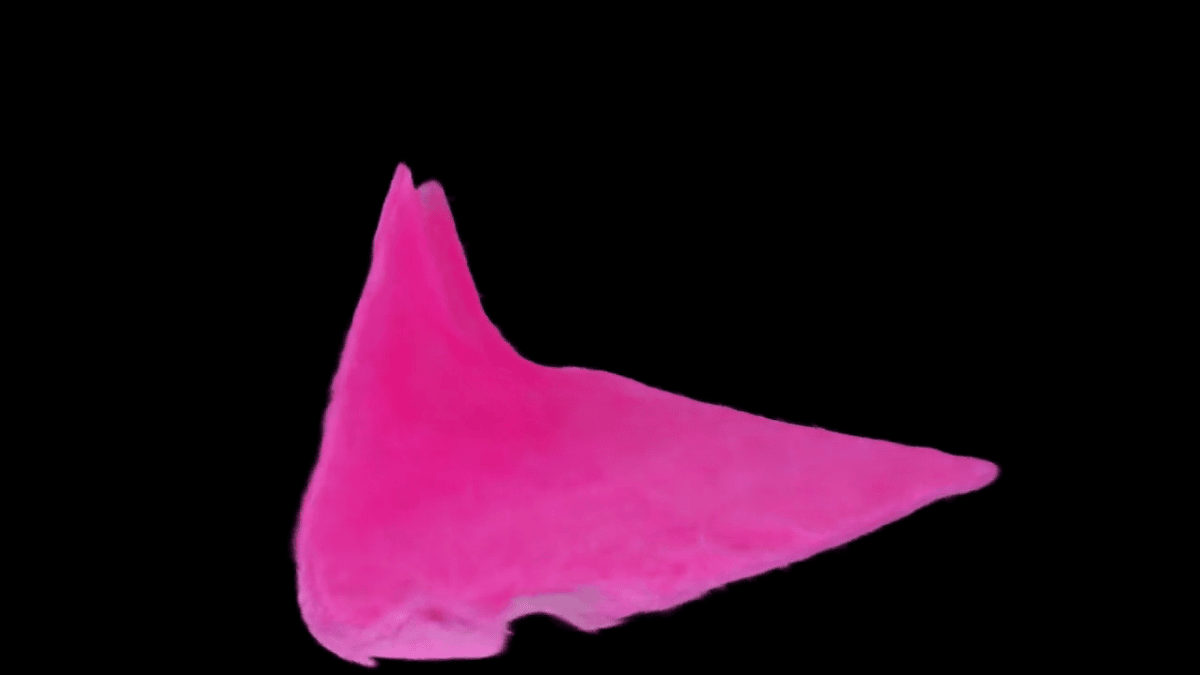}
        \end{subfigure}
        \hfill
        \begin{subfigure}{0.24\linewidth}
            \centering
            \includegraphics[width=\linewidth]{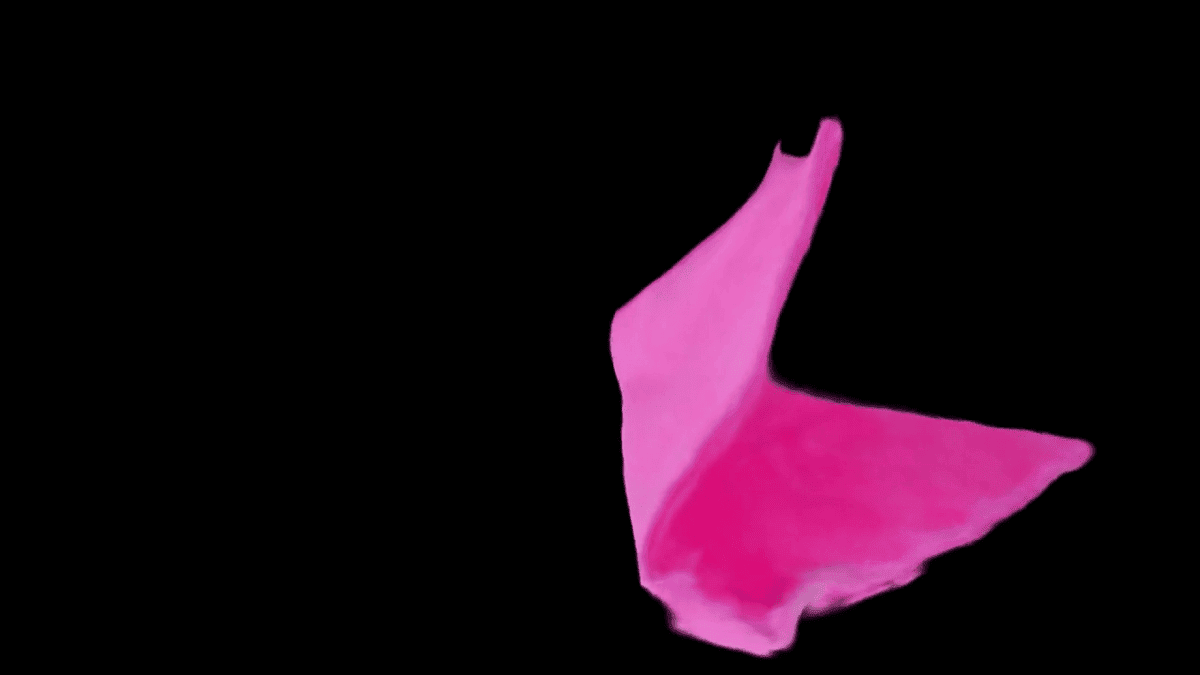}
        \end{subfigure}
        \hfill
        \begin{subfigure}{0.24\linewidth}
            \centering
            \includegraphics[width=\linewidth]{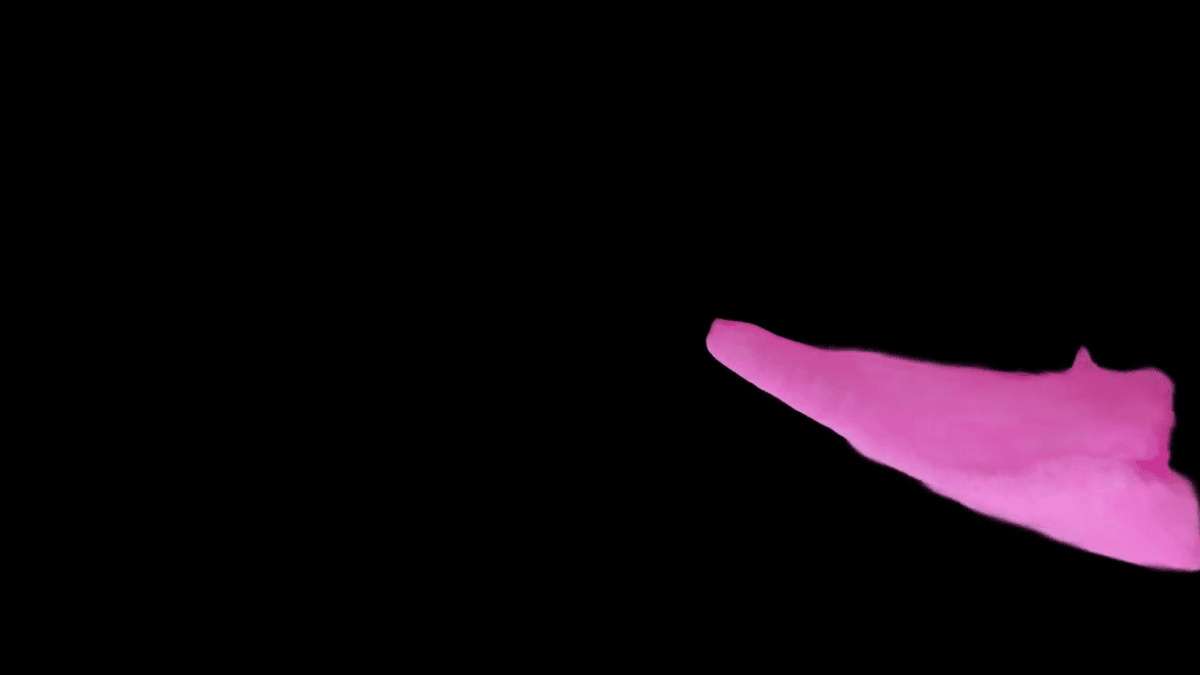}
        \end{subfigure}

    \end{minipage}
    \caption{Visualization of dynamic reconstruction results from novel viewpoints and different temporal steps.}
    \label{fig:reconstruction}
\end{figure*}

\begin{figure*}[t!]
    \centering
    \includegraphics[width=\linewidth]{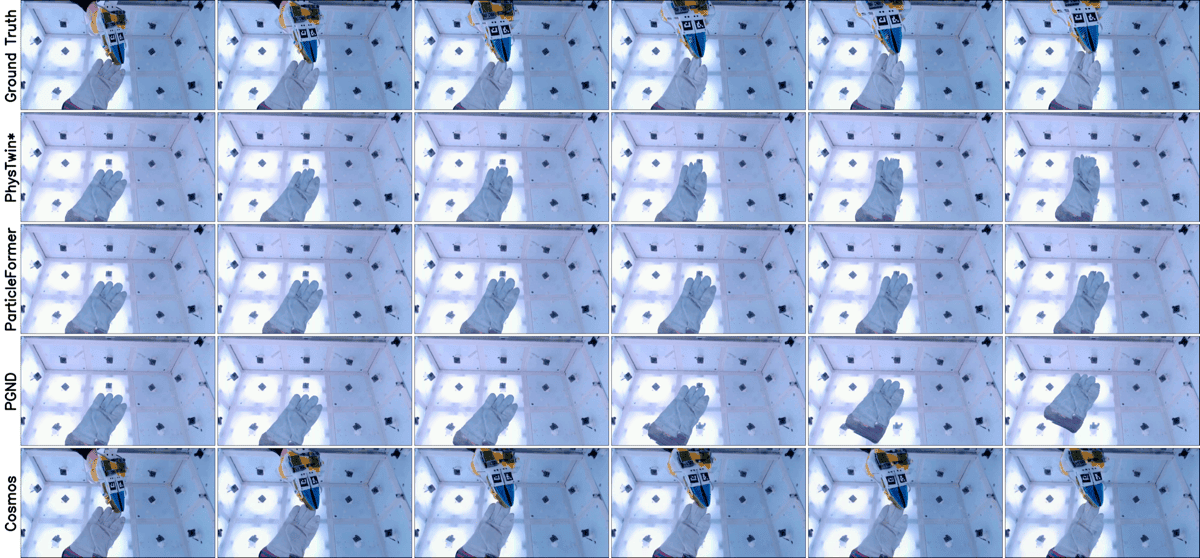} \\
    \vspace{2pt}
    \includegraphics[width=\linewidth]{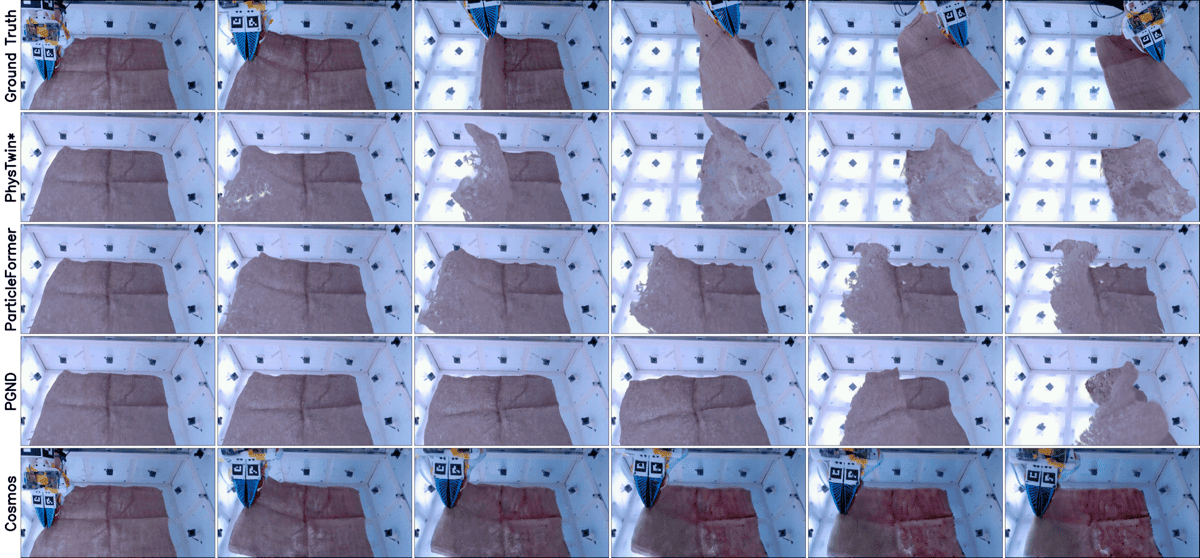}
    \caption{Visualization of Multi-Episode Generalization. We show the predicted future frames for the glove-cloth (top) and sack-cloth (bottom) objects, comparing different world models under the episode generalization setting.}
    \label{fig:episode_generalization}
\end{figure*}

\section{Multi-Episode Generalization Visualization}
\label{sec:episode_generalization_details}
In \Cref{fig:episode_generalization}, we provide a detailed visualization of the multi-episode generalization capabilities of different world models on unseen interactions involving the glove and sack objects. 
The visualization compares the ground truth (GT) sequence against the action-conditioned predictions from Cosmos, ParticleFormer, PGND, and PhysTwin. 

Note that the entry labeled \textbf{PhysTwin*} in the figure indicates a special case. 
Since PhysTwin is strictly an optimization-based physical simulator that requires system identification to fit physical parameters directly on an observed trajectory, it cannot directly ``predict'' an entirely unseen episode zero-shot in the same manner as the forward-pass neural models (Cosmos, ParticleFormer, and PGND). 
Therefore, for PhysTwin*, we allow the optimization process to observe $80\%$ of the transitions in this specific testing episode to fit its parameters, and then simulate the subsequent steps. 
We include its results here strictly as an upper-bound reference for the performance of a fully optimized physics simulator under partial observation, rather than as a direct comparison of zero-shot episode generalization.

\section{Dataset Object Taxonomy and Diversity}
The \shortname dataset is designed to capture the vast complexity of real-world deformable object dynamics. 
We categorize the 198 daily-life objects into three primary classes based on their topological dimensionality and physical response: (1) 28 1D deformables; (2) 98 2D deformables; and (3) 72 3D volumetric deformables. 
These objects further span 17 everyday semantic categories with a naturally long-tailed distribution.
A comprehensive comparison of the \shortname dataset against existing benchmarks, including a full list of supported manipulation actions and object categories, is provided in Table~\ref{table:dataset_full}.
This taxonomy facilitates research into category-specific dynamics and cross-category generalization.

\subsection{1D Deformables}
1D deformable objects are characterized by their linear structure, where one dimension significantly exceeds the others. 
Our collection includes a wide variety of ropes, cables, and wires, spanning a broad spectrum of material stiffness, thicknesses, and surface textures. 
These objects exhibit complex behaviors such as knotting, entangling, and significant self-occlusion during manipulation. 
To further differentiate their physical properties, we divide them into three subcategories:
\begin{itemize}
    \item \textbf{Thick rope-like deformables}: Objects with relatively high bending stiffness and substantial diameter, such as cotton ropes, climbing ropes, and belts.
    \item \textbf{Thin rope-like deformables}: Highly flexible objects with small diameters, including threads, silk ribbons, and paracords.
    \item \textbf{Stiff cable-like deformables}: Objects with significant elastic memory and higher torsional stiffness, such as USB cables, garden hoses, and metal chains.
\end{itemize}
The specific objects collected are detailed in Table~\ref{table:1d deformables} and visualized in Fig.~\ref{fig:1d_deformables}.

\begin{table}[t!]
\renewcommand{\arraystretch}{1.5}
\setlength{\tabcolsep}{6pt}

\centering
\resizebox{\textwidth}{!}{%

\begin{tabular}{p{2.5cm}p{9cm}}
\toprule
 & \textbf{1D Deformables} \\
\midrule
\textbf{Thick rope-like deformables}        & cotton rope, rubber band, long thin fur, belt, watch strap, nylon rope, jump rope, climbing rope, cotton clothesline, twisted hemp rope, flat strip, silicone straw \\
\textbf{Thin rope-like deformables}      & silk ribbon, thread, feather, paracord, shoelace, leather strip, rubber band, fishing line, yarn, pipe cleaner \\
\textbf{Stiff cable-like deformabels}  & USB cable, thin chain, beaded string, metal chain, alloy, garden hose, medical tube \\
\bottomrule
\end{tabular}
}
\vspace{4pt}

\caption{\textbf{Detailed Categorization for 1D Deformables.} The 1D deformable objects collected in \shortname can be divided into thick rope-like deformables, thin rope-like deformables, and stiff cable-like deformables.}
\label{table:1d deformables}
\end{table}

\subsection{2D Deformables}
2D deformables, or thin-shell objects, represent a common yet challenging class in robotics. 
Our dataset includes diverse fabrics, cloths, garments, and paper-like materials that exhibit complex folding, wrinkling, and multi-modal contact dynamics. 
Our dataset contains diverse material properties, ranging from the highly compliant silk to the relatively stiff paper and plastic.
This diversity allows for testing a model's ability to capture varying degrees of elasticity and plasticity. 
We categorize these into:
\begin{itemize}
    \item \textbf{Cloths and Fabrics}: Daily-use textiles like table cloths, towels, and handkerchiefs with various textures and weights.
    \item \textbf{Garments}: Complex geometric structures such as shirts, socks, hats, and gloves, which pose significant challenges for tracking and state estimation.
    \item \textbf{Bag-like and Paper-like deformables}: Objects that exhibit significant plastic deformation or non-linear elastic responses, such as airbags, trash bags, and various types of paper.
\end{itemize}
The categorized instances are listed in Table~\ref{table:2d deformables} and visualized across Fig.~\ref{fig:2d_deformables_1}, Fig.~\ref{fig:2d_deformables_2}, and Fig.~\ref{fig:2d_deformables_3}.

\begin{table}[t!]
\renewcommand{\arraystretch}{1.5}
\setlength{\tabcolsep}{6pt}

\centering
\resizebox{\textwidth}{!}{%

\begin{tabular}{p{2cm}p{9cm}}
\toprule
 & \textbf{2D Deformables} \\
\midrule
\textbf{Cloths}        &  pink table cloth, yellow table cloth, orange table cloth, green table cloth, chessboard-textured cloth, handkerchief, cleaning cloth, glass cleaner\\
\textbf{Fabrics}        & pocket bag, small grey pocket bag, black pocket bag, small mat, yellow glove, green glove, gray glove, pencil case, towel, curtain, blanket, apron, felt sheets, tulle fabric, bottle cover, net \\
\textbf{Garments}        &  cotton hat, stiff glove, shirt, sock, necktie, kneepad, cotton half gloves, beret hat, silk scarf, cotton scarf, head band, wrist band, hiking glove \\
\textbf{Paper-like deformables}        & wipe, napkin, paper, wrap paper, napkin case, poster paper, envelope, crepe paper, origami paper, shredded packing paper, parchment paper, sticker paper, fake lettuce, paper towel, rubber jar opener  \\
\textbf{Bag-like deformables}        & airbag, plastic trash bag, transparent trash bag, umbrella bag, ziplog bag, plastic takeout bag, bakery bag, vaccuum-sealed bag, mesh produce bag, sack, jewelry pouch, purse \\
\textbf{Other thin-shell deformables}        & cutting mat, vinyl glove, thin foam, flat white foam, bandage, cotton pad, teabag, mask, mop, candy packet, bottle strap, plastic sheet, oven mitt, drying mat, pouch, non-transparent shower cap, transparent shower cap, cotton gauze, flat roller foam, foam alphabet letters, baking mat, finger wrap, hydrogel patch,  bubble wrap, fake banana peel, fake flower, small box, big box, whole wheat bread, pita bread, inflatable beach ball cover, hand puppet, gel pack coaster, rubber glove, shoe sole, silicone wrist band \\
\bottomrule
\end{tabular}
}
\vspace{4pt}

\caption{\textbf{Detailed Categorization for 2D Deformables.} The 2D deformable objects collected in \shortname can be divided into cloths, fabric, garments, paper-like deformables, bag-like deformables, and other thin-shell deformables.}
\label{table:2d deformables}
\end{table}

\subsection{3D Volumetric Deformables}
3D volumetric deformables represent objects with non-negligible thickness and complex internal structures. 
This category includes plush toys, stuffed animals, and various foam-based objects that exhibit significant volume-preserving or volume-changing shape changes during interaction. 
Capturing their dynamics requires modeling internal stress and long-range dependencies within the object volume. 
Our collection is divided into:
\begin{itemize}
    \item \textbf{Stuffed Animals and Plush Toys}: Objects with highly complex shapes and internal padding, such as teddy bears and various animal toys.
    \item \textbf{Foam and Squeezables}: Materials with varying density and recovery rates, including sponges, stress balls, and household items like hand sanitizers and shoes.
\end{itemize}
The categorized instances are listed in Table~\ref{table:3d deformables} and visualized in Fig.~\ref{fig:3d_deformables_1}, Fig.~\ref{fig:3d_deformables_2}, and Fig.~\ref{fig:3d_deformables_3}.

\begin{table}[t!]
\renewcommand{\arraystretch}{1.5}
\setlength{\tabcolsep}{6pt}

\centering
\resizebox{\textwidth}{!}{%

\begin{tabular}{p{2cm}p{9cm}}
\toprule
 & \textbf{3D Deformables} \\
\midrule
\textbf{Stuffed Animals}   & dog, doll, cartoon cat, big cat, spider, penguin, monster, polar bear, brown bear, monkey, green sloth, grey sloth, frog, blue fish, orange fish, brown hello kitty, white hello kitty, snake, rabbit, jellyfish, teddy bear, sheap \\
\textbf{Plush Toys} & ball, boxing target, cube, rubber duck, plush football, squeezable squirrel, squeezable fruit, beanbag, octopus, stress ball, bread plush, croissant plush\\
\textbf{Foam} & sponge, soft rectangle sponge, butter sponge, purple makeup sponge, pink makeup sponge, big kitchen sponge, fake cheese, small foam roll, ear plug, kitchen foam, sponge stamp, thick foam \\
\textbf{Squeezables} & cream squeezer, toothpaste, napkin roll, hand sanitizer, shoe, cup, birthday hat, baby ring toy, pillow, silicone teething toy, stacking cups, clay, fake ice pack, cervical collar, pen grip, stress cube, stress donut, animal toy, rubber ball, eraser, squishy dog, squisy frog, baking mold, slime, cupcake toy, crystal slime\\
\bottomrule
\end{tabular}
}
\vspace{4pt}

\caption{\textbf{Detailed Categorization for 3D Deformables.} The 3D deformable objects collected in \shortname can be divided into stuffed animals, plush toys, foam, and squeezables.}
\label{table:3d deformables}
\end{table}

\section{Dataset Visualizations}
\label{sec:dataset_visualizations}
\begin{figure*}
    \centering
    \includegraphics[width=\linewidth]{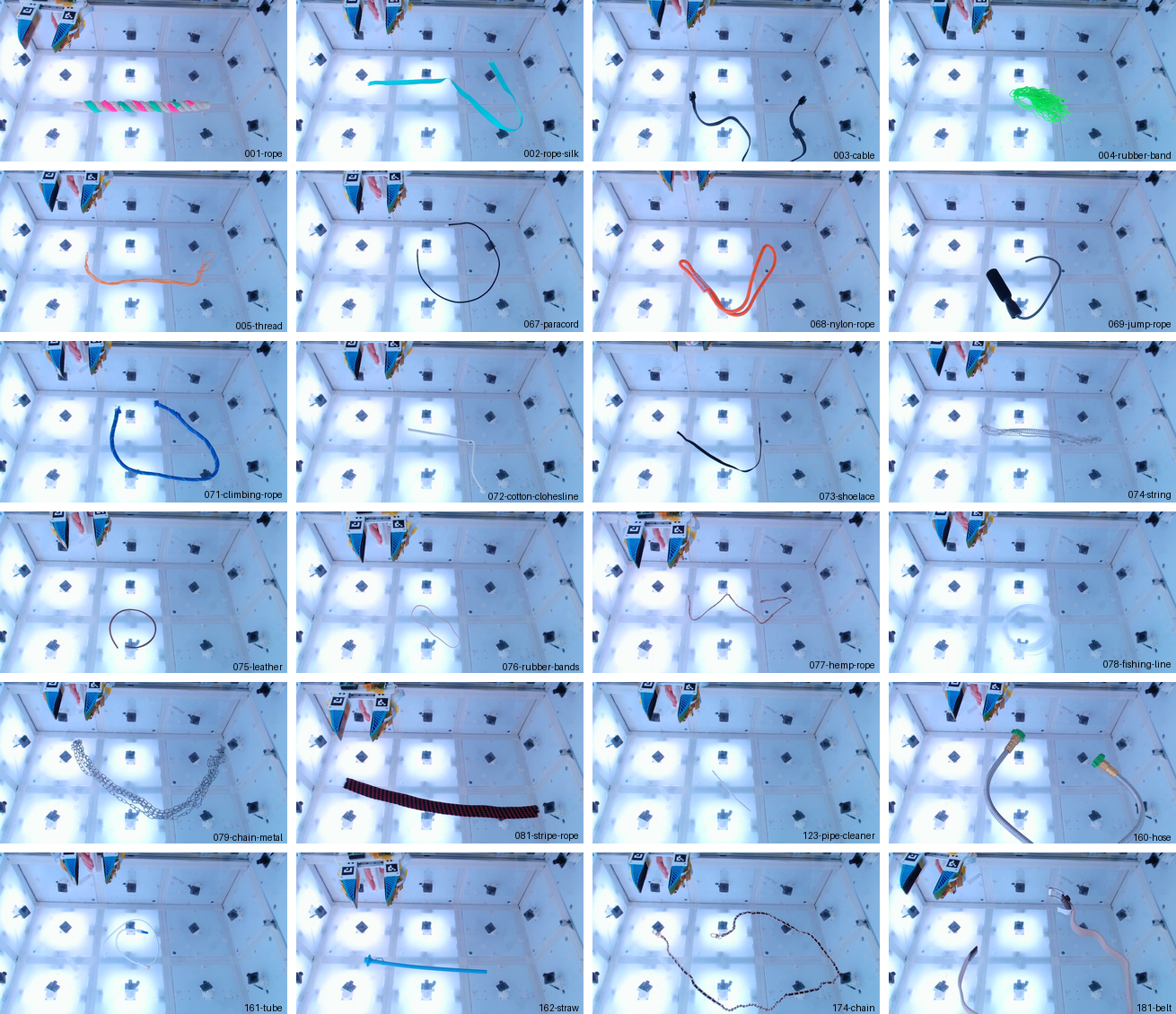}
    \caption{\textbf{Visualization of 1D Deformables}. Object names are marked at the lower right corner of each image.}
    \label{fig:1d_deformables}
\end{figure*}

\begin{figure*}
    \centering
    \includegraphics[width=\linewidth]{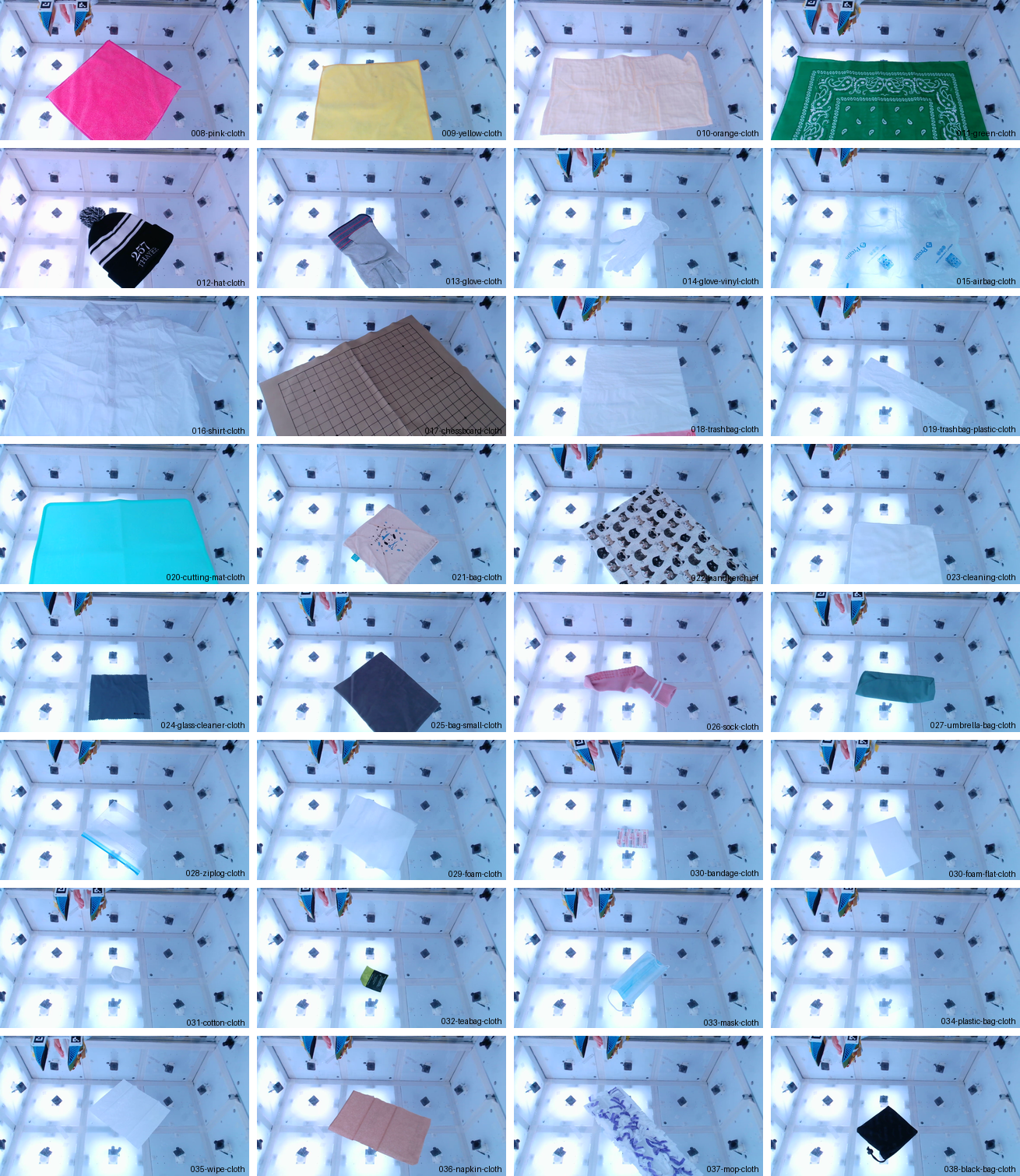}
    \caption{\textbf{Visualization of 2D Deformables}. Object names are marked at the lower right corner of each image.}
    \label{fig:2d_deformables_1}
\end{figure*}

\begin{figure*}
    \centering
    \includegraphics[width=\linewidth]{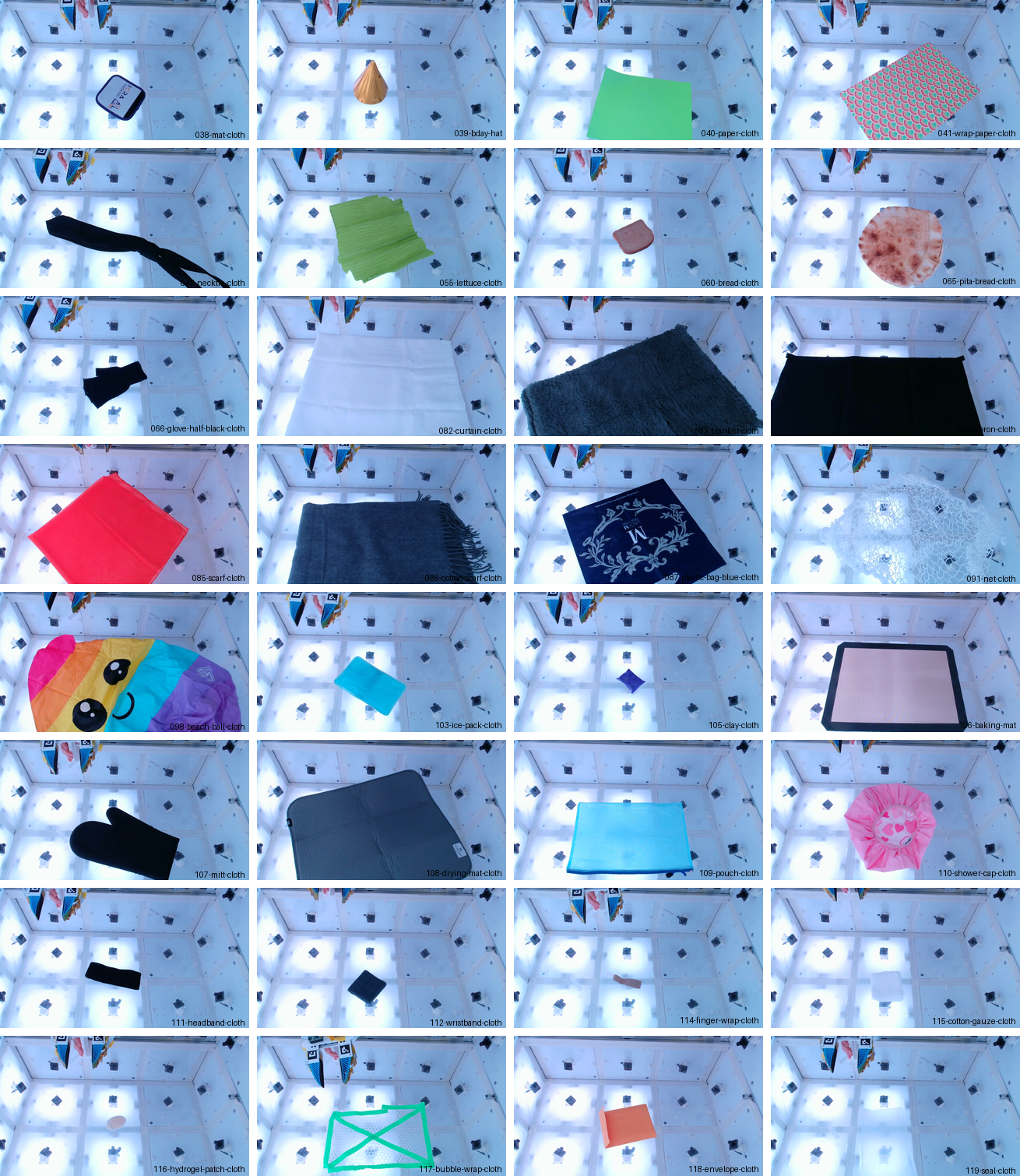}
    \caption{\textbf{Visualization of 2D Deformables (continued)}. Object names are marked at the lower right corner of each image.}
    \label{fig:2d_deformables_2}
\end{figure*}

\begin{figure*}
    \centering
    \includegraphics[width=\linewidth]{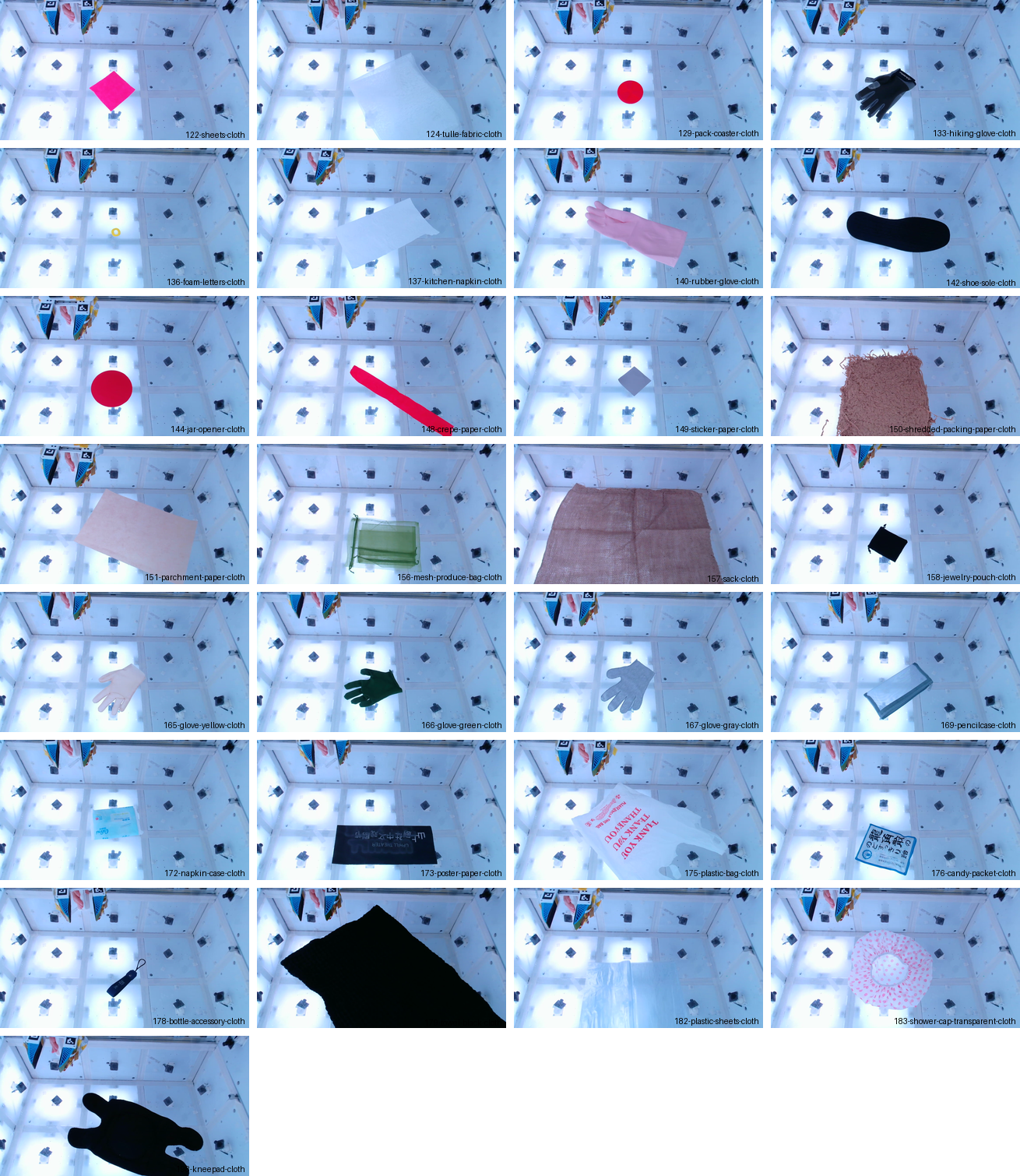}
    \caption{\textbf{Visualization of 2D Deformables (continued)}. Object names are marked at the lower right corner of each image.}
    \label{fig:2d_deformables_3}
\end{figure*}

\begin{figure*}
    \centering
    \includegraphics[width=\linewidth]{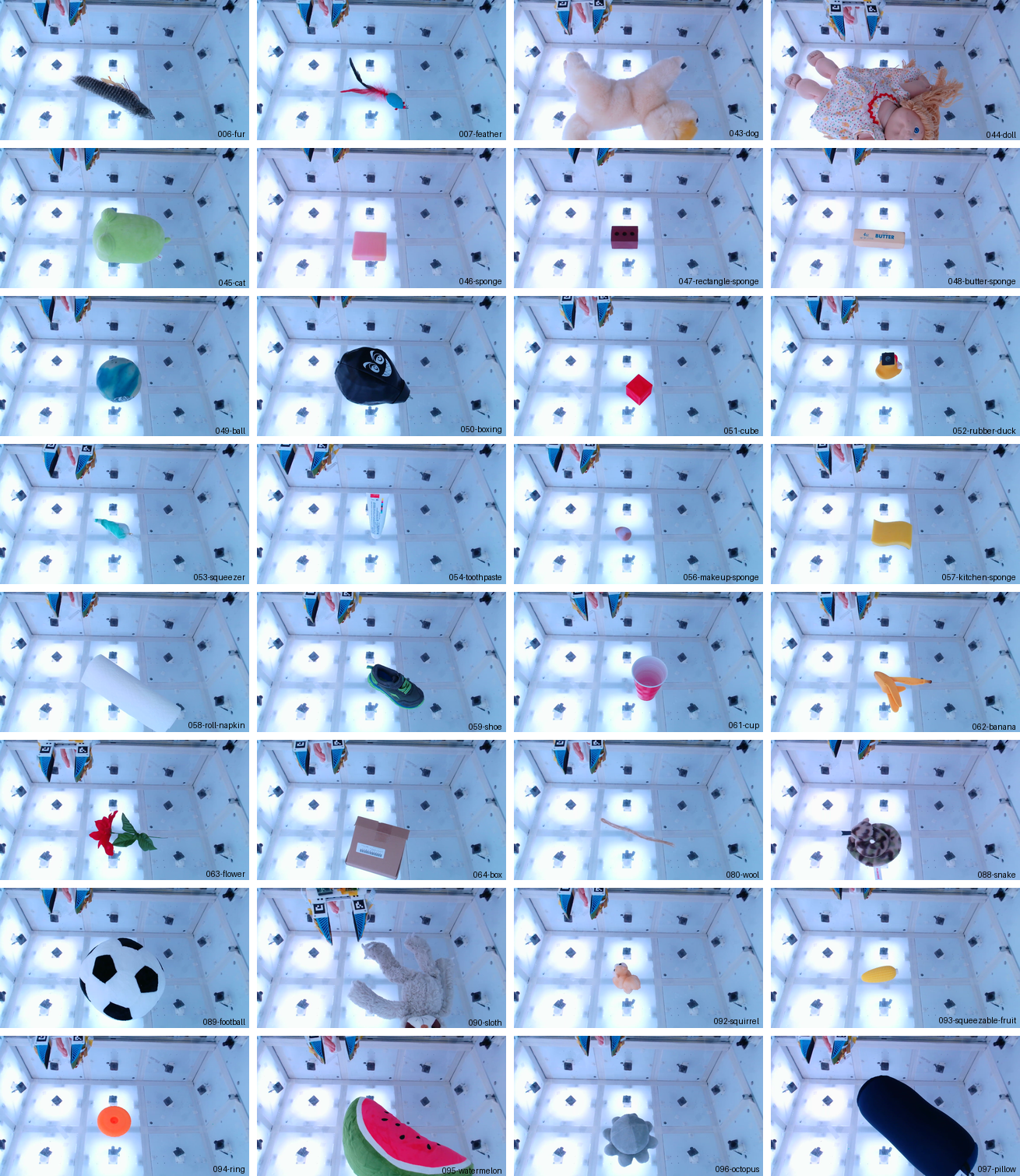}
    \caption{\textbf{Visualization of 3D Deformables}. Object names are marked at the lower right corner of each image.}
    \label{fig:3d_deformables_1}
\end{figure*}

\begin{figure*}
    \centering
    \includegraphics[width=\linewidth]{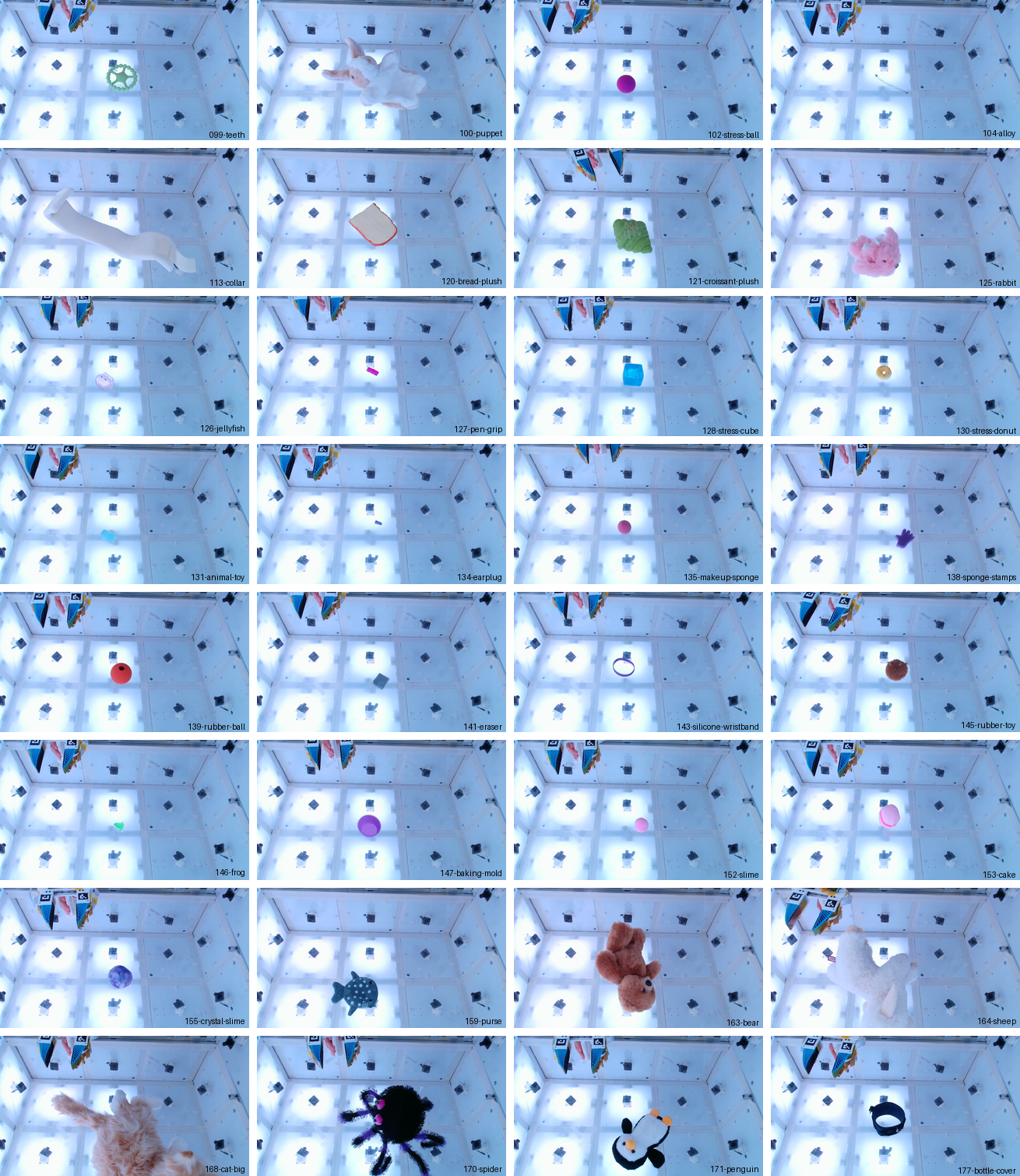}
    \caption{\textbf{Visualization of 3D Deformables (continued)}. Object names are marked at the lower right corner of each image.}
    \label{fig:3d_deformables_2}
\end{figure*}

\begin{figure*}
    \centering
    \includegraphics[width=\linewidth]{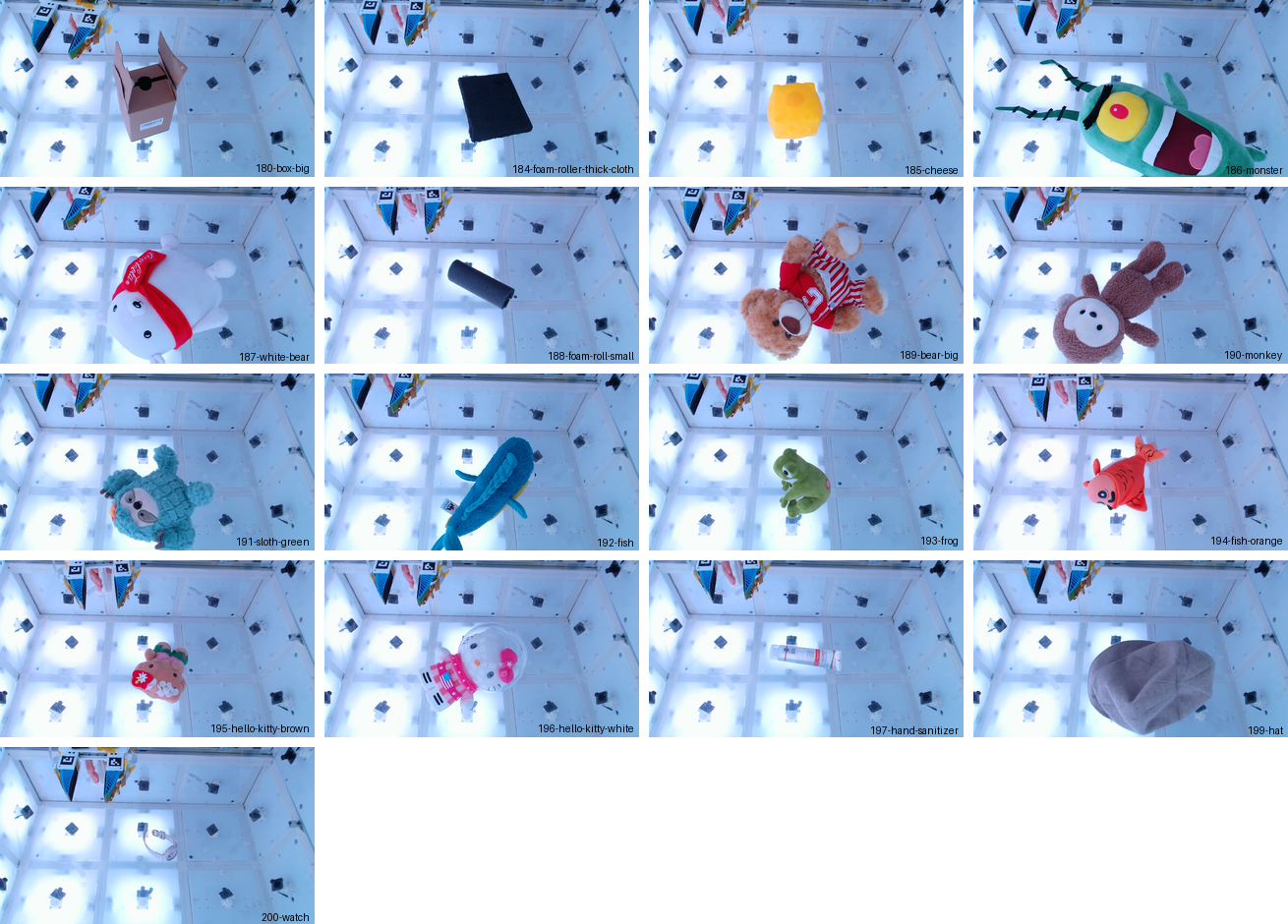}
    \caption{\textbf{Visualization of 3D Deformables (continued)}. Object names are marked at the lower right corner of each image.}
    \label{fig:3d_deformables_3}
\end{figure*}

\begin{figure*}
    \centering
    \includegraphics[width=\linewidth]{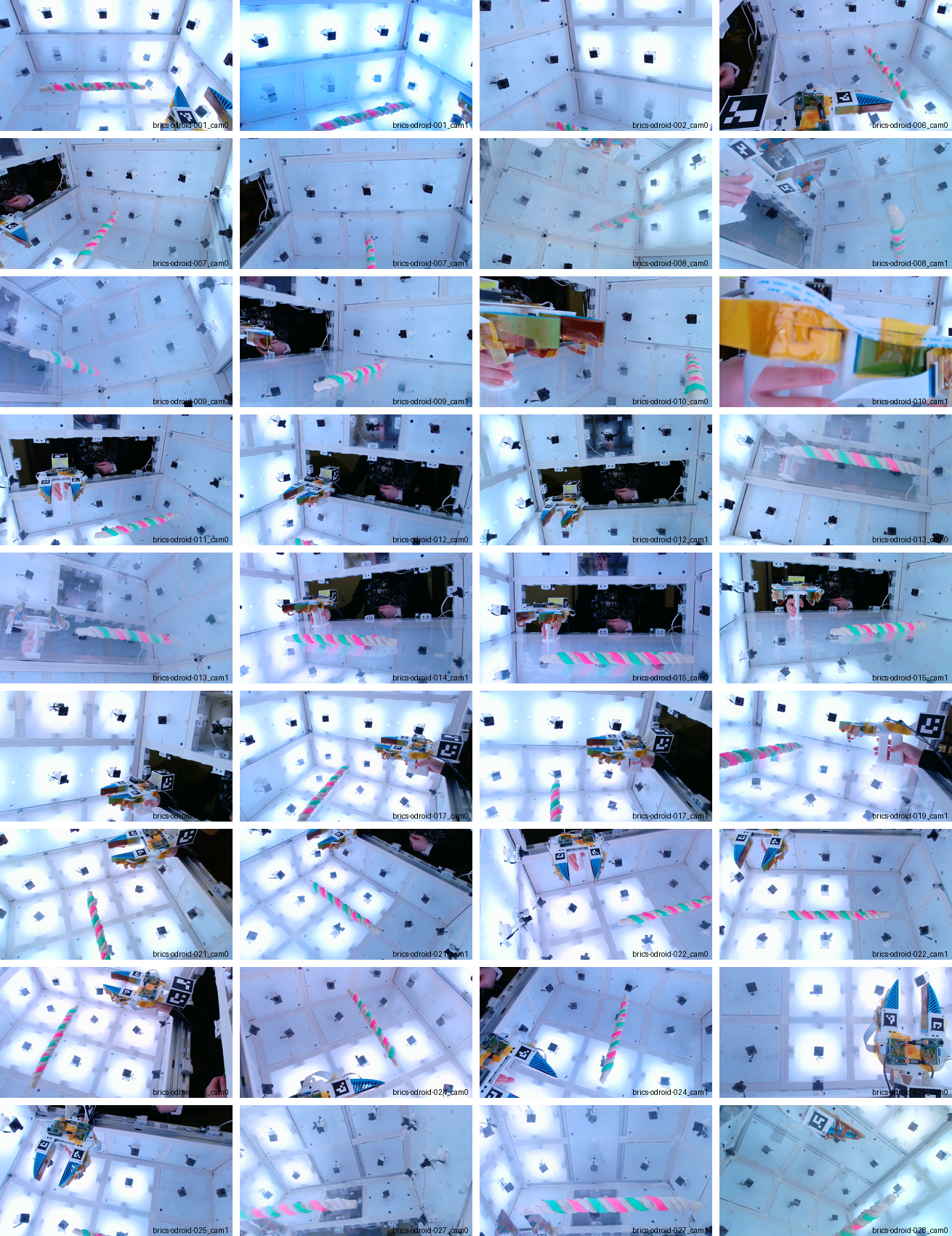}
    \caption{\textbf{Visualization of Multi-view Deformables}. Each image represents a different view of the same object.}
    \label{fig:multi-view_deformables_001_rope}
\end{figure*}

\begin{figure*}
    \centering
    \includegraphics[width=\linewidth]{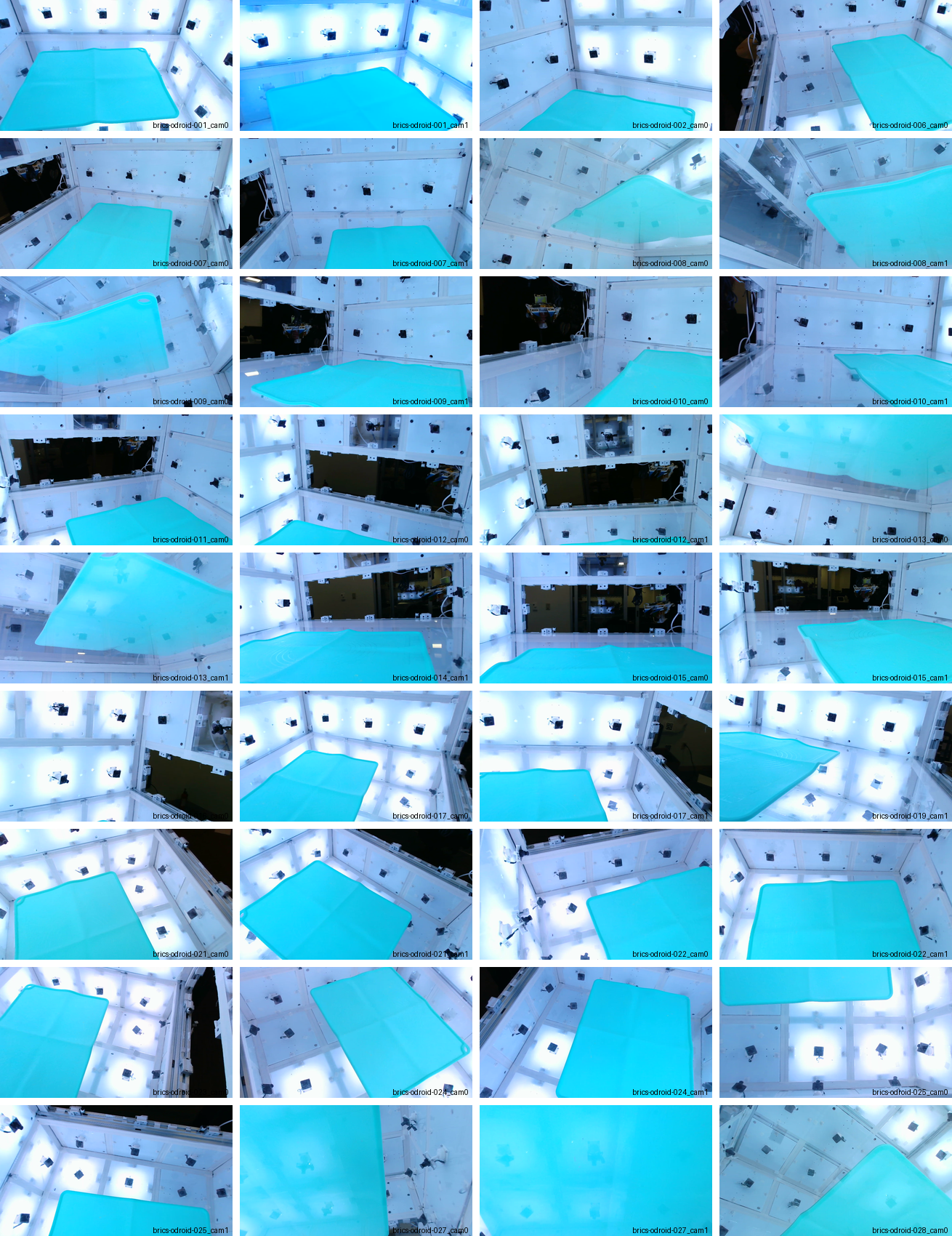}
    \caption{\textbf{Visualization of Multi-view Deformables}. Each image represents a different view of the same object.}
    \label{fig:multi-view_deformables_020_cutting-mat-cloth}
\end{figure*}

\begin{figure*}
    \centering
    \includegraphics[width=\linewidth]{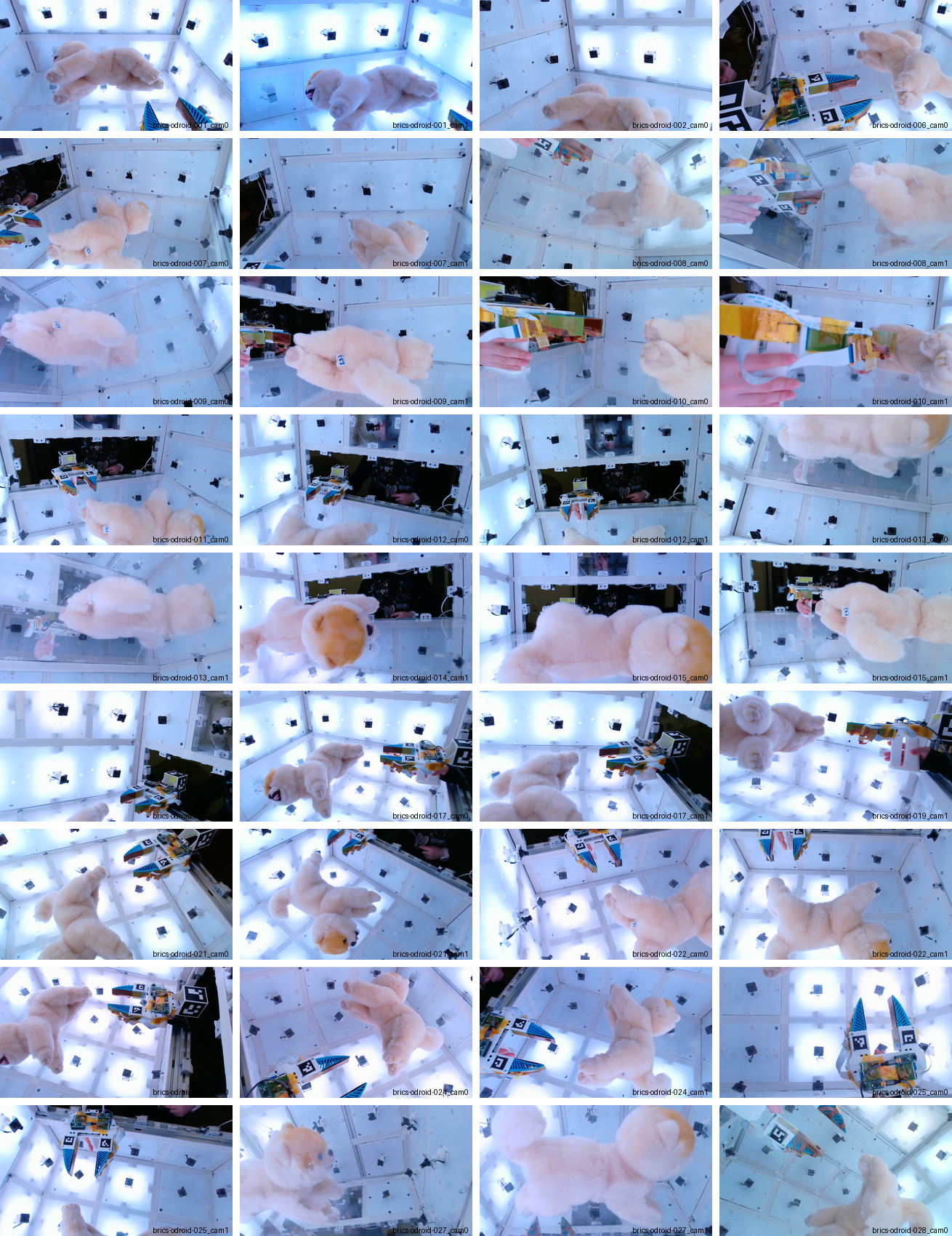}
    \caption{\textbf{Visualization of Multi-view Deformables}. Each image represents a different view of the same object.}
    \label{fig:multi-view_deformables_043_dog}
\end{figure*}

To provide a comprehensive view of the object diversity and our multi-view capture setup, we include an extensive set of visualizations.
Figures~\ref{fig:1d_deformables} through \ref{fig:3d_deformables_3} showcase single-view images of the diverse 1D, 2D, and 3D volumetric deformable objects collected in the \shortname dataset. 
Furthermore, Figures~\ref{fig:multi-view_deformables_001_rope} through \ref{fig:multi-view_deformables_043_dog} demonstrate the synchronized multi-view perspectives (from our 41-camera setup) for selected objects, illustrating how the dense camera array captures complex interactions and mitigates self-occlusion from multiple distinct angles.

\end{document}